\documentclass[11pt]{article}

\usepackage[final]{acl}
\usepackage{enumitem}
\usepackage{times}
\usepackage{latexsym}
\usepackage{algorithm}
\usepackage[T1]{fontenc}
\usepackage{amsmath}
\usepackage[most]{tcolorbox}
\usepackage[ruled,vlined,algo2e]{algorithm2e}
\usepackage{amssymb}
\usepackage{caption}
\usepackage{booktabs}
\usepackage{multirow}
\usepackage{array}
\usepackage{xcolor}
\usepackage{colortbl}
\usepackage{makecell}

\usepackage{xfp}      
\usepackage{siunitx}  
\usepackage{pifont}   
\newtcolorbox{promptbox}[1][]{
  enhanced,      
  breakable,       
  colback=gray!5,
  colframe=gray!50!black,
  title=\textbf{#1},
  fonttitle=\bfseries,
  sharp corners,
  boxrule=0.5mm,
  left=2mm, right=2mm, top=2mm, bottom=2mm,
  fontupper=\small 
}
\usepackage[utf8]{inputenc}
\usepackage{tabularx} 
\usepackage{microtype}

\usepackage{inconsolata}
\usepackage{algorithmic}
\usepackage{graphicx}

\title{\textbf{UniGeM}: \underline{Uni}fying Data Mixing and Selection \\ via \underline{G}eometric \underline{E}xploration and \underline{M}ining}

\author{
  Changhao Wang\thanks{Equal contribution} \\
  Politecnico di Torino \\
  \texttt{changhao.wang@polito.it}
  \And
  Yunfei Yu\footnotemark[1] \\
  Ant Group
  \AND
  Xinhao Yao \\
  Renmin University of China
  \And
  Jiaolong Yang \\
  Ant Group
  \And
  Riccardo Cantoro \\
  Politecnico di Torino
  \AND
  Chaobo Li \\
  Institute of Microelectronics of the CAS
  \And
  Qing Cui \\
  Ant Group
  \And
  Jun Zhou \\
  Ant Group
}

\begin{document}
\maketitle
\begin{abstract}
The scaling of Large Language Models (LLMs) is increasingly limited by data quality. Most methods handle data mixing and sample selection separately, which can break the structure in code corpora. We introduce \textbf{UniGeM}, a framework that unifies mixing and selection by treating data curation as a \textit{manifold approximation} problem without training proxy models or relying on external reference datasets. UniGeM operates hierarchically: \textbf{Macro-Exploration} learns mixing weights with stability-based clustering; \textbf{Micro-Mining} filters high-quality instances by their geometric distribution to ensure logical consistency. Validated by training 8B and 16B MoE models on 100B tokens, UniGeM achieves \textbf{2.0$\times$ data efficiency} over a random baseline and further improves overall performance compared to SOTA methods in reasoning-heavy evaluations and multilingual generalization.

\end{abstract}

\section{Introduction}

The generalization of Large Language Models (LLMs) has traditionally relied on scaling parameters and data volume \citep{kaplan2020scaling}. 
However, recent shifts in scaling laws suggest that data quality now constrains model performance more than raw quantity \citep{hoffmann2022chinchilla, gadre2024language}. 
As high-quality public corpora are nearing depletion \citep{villalobos2022will}, simply aggregating noisy web data yields diminishing returns \citep{gunasekar2023textbooks}.
This motivates a focus on \textbf{data efficiency}: finding subsets that give more gain per compute \citep{li2023starcoder, sorscher2022beyond}.

\begin{figure}[t]
\centering
 \includegraphics[width=1\linewidth]{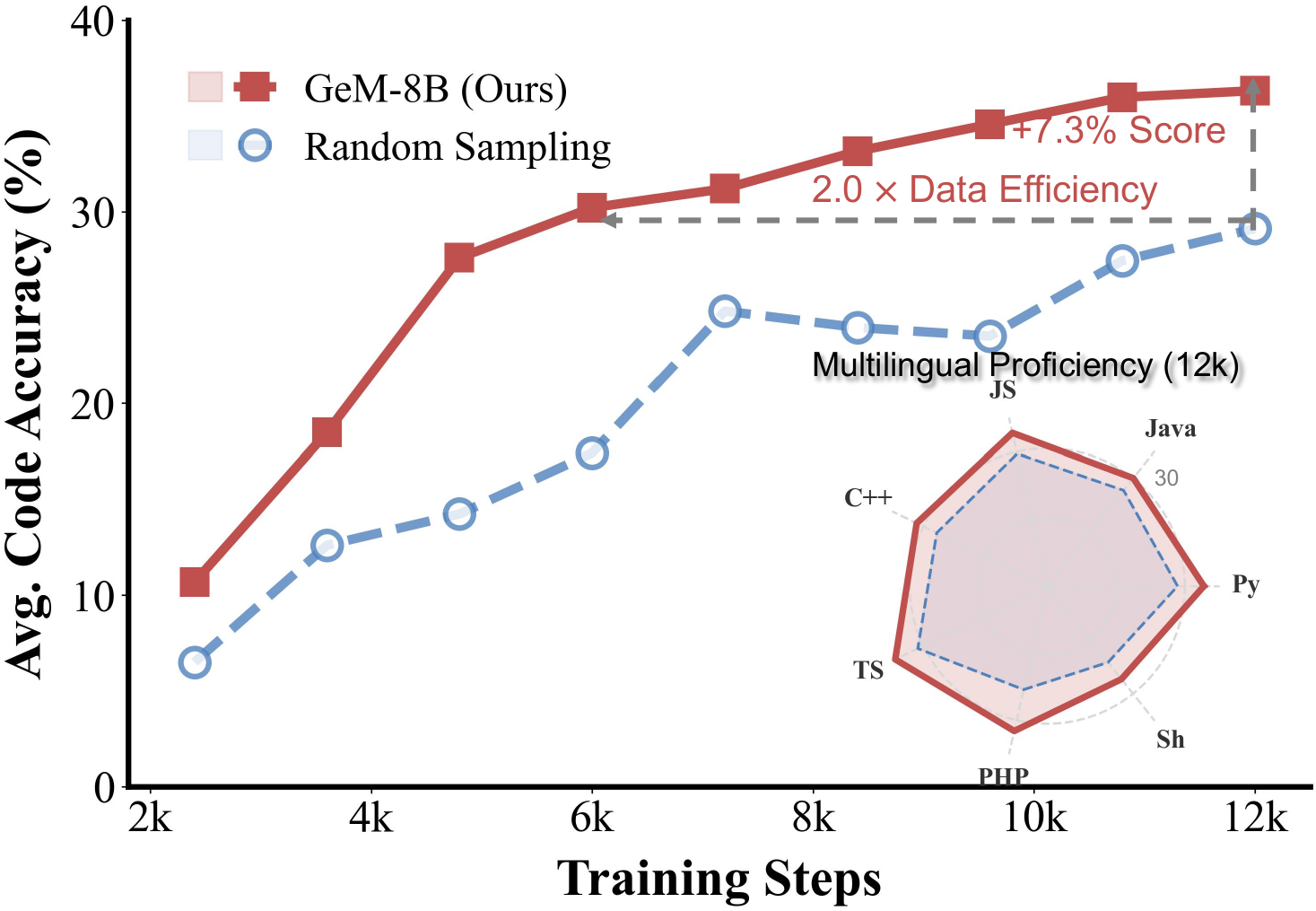}
    \caption{\textbf{Average downstream performance:} Random sampling vs.\ UniGeM for pre-training an 8B MoE model. The inset radar chart shows that UniGeM achieves stronger multilingual performance across programming languages.}
 \label{fig:teaser}
\end{figure}

Most data efficiency approaches fall into two buckets: domain mixing and instance selection~\citep{xie2023doremi,yu2024mates,liu2025quadmix}. 
Data mixing serves as a \textbf{macro-distribution balancing} mechanism~\citep{chen2023skill}, but its coarse granularity often treats domains as flat distributions and overlooks structure within each domain~\citep{diao2025climb}. 
Meanwhile, many selection methods require training proxy models or reference datasets, which adds substantial overhead and may not track the target model's behavior at scale~\citep{mass,yu2024mates,li2024scalingfilter}. 
Other selection methods, including heuristics and LLM-based scoring, evaluate samples independently and ignore the underlying manifold structure~\citep{zhuang2025metarater,xie2023dsir}.
This decoupling leads to structural blind spots. We either optimize macro-distribution without considering micro-quality, or filter samples at the cost of disrupting the hierarchical dependencies.
This matters most for the \textbf{code corpus}, a rigid logical manifold defined by brittle syntax and hierarchical dependencies \citep{li2023starcoder,guo2020graphcodebert,feng2020codebert}. 
We thus ask: \textit{How can we unify macro-distribution balancing and micro-quality selection to identify the "golden subset" within a structured code corpus?}

To bridge this gap, we introduce \textbf{UniGeM (Geometric Exploration and Mining)}, a framework that unifies macro-distribution balancing and micro-quality selection. Unlike existing methods that rely on external reference datasets for alignment or require training expensive proxy models to estimate data importance, UniGeM views these tasks as a unified \textit{manifold approximation} problem. 
First, \textbf{Macro-Exploration (Stage-I)} discovers semantic manifolds via stability-driven clustering to optimize macro-distribution balancing. 
Subsequently, \textbf{Micro-Mining (Stage-II)} performs micro-quality selection using intrinsic geometric priors to capture structural and logical dependencies. 
By measuring manifold deviations, UniGeM distills a \textbf{golden subset} that serves as a faithful approximation of the data manifold, preserving global topology and local dependency structure.
We validate UniGeM by training 8B and 16B Mixture-of-Experts (MoE) models (both with 1.4B active parameters) from scratch on a 100B-token code-and-text mixture.
Results show that UniGeM achieves superior data efficiency and model performance compared to existing baselines.
Our main contributions are:
\begin{itemize}[leftmargin=*, noitemsep, topsep=0pt]
    \item \textbf{Unified Framework:} We propose a geometry-centric framework to unify macro-distribution balancing and micro-quality selection via manifold approximation.
    
    \item \textbf{Proven Data Efficiency:} UniGeM achieves \textbf{$2.0\times$ data efficiency} (vs.\ a random baseline) and performs better after one epoch.
    
    \item \textbf{Broader Gains:} UniGeM improves overall performance over SOTA baselines and shows stronger multilingual generalization.

    \item \textbf{Topological Insights:} Our ablations suggest that coverage and local structure matter for reasoning beyond per-sample quality scores, consistent with a manifold-approximation view of curation.
        
\end{itemize}
\section{Proposed Method: The UniGeM Framework}
\label{sec:method}

In this section, we detail the implementation of UniGeM and provide a theoretical analysis to demonstrate effectiveness in data curation.

\begin{figure*}[t]
    \centering
    \includegraphics[width=1\linewidth]{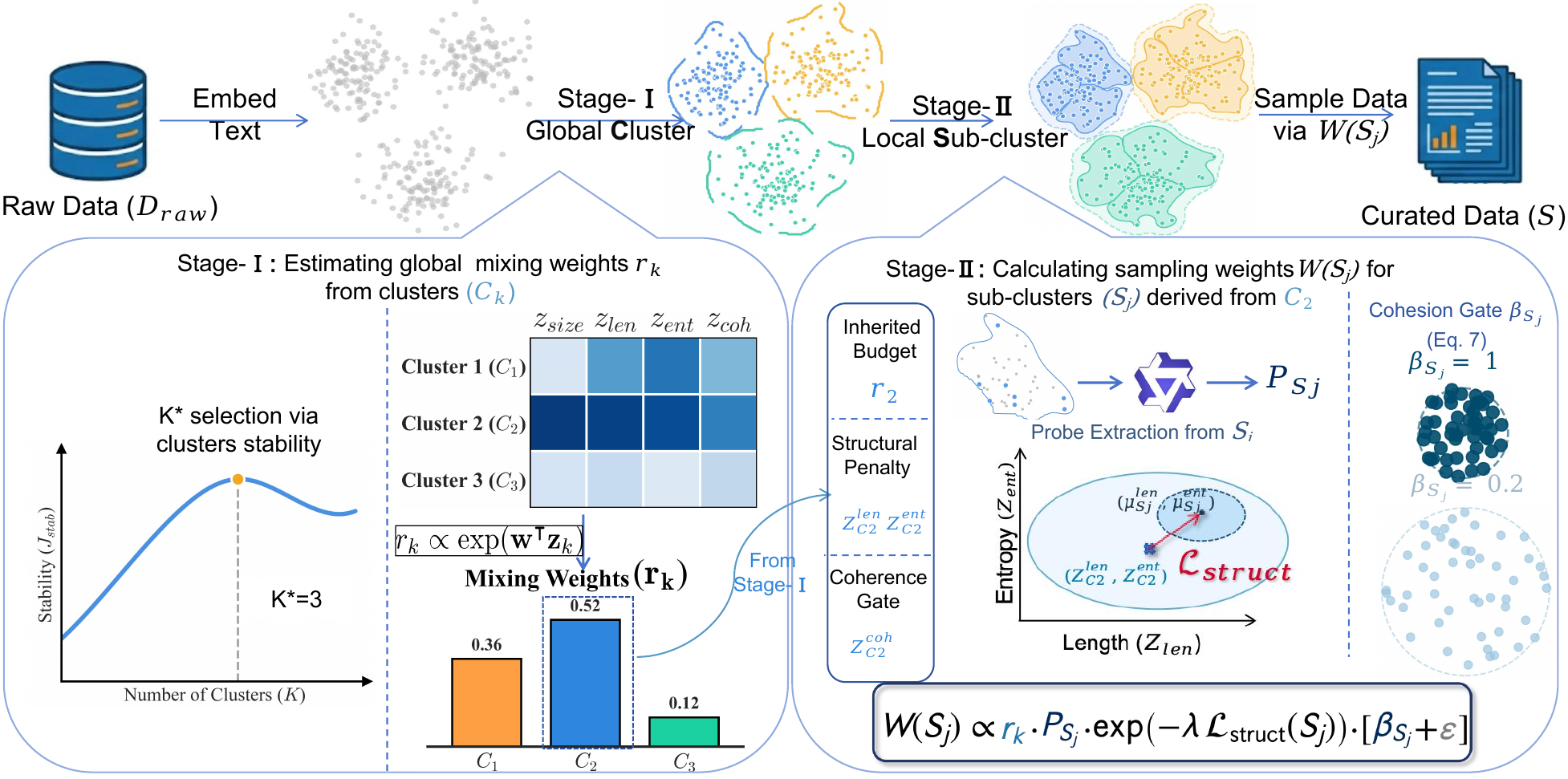} 
\caption{\textbf{Overview of the UniGeM framework.} The curation pipeline operates in two hierarchical stages: 
(Left) \textbf{Stage-I (Macro-Exploration)} identifies the intrinsic manifold resolution $K^*$ via topological stability (Algorithm~\ref{alg:intrinsic_resolution}) and derives global mixing weights $\mathbf{r}$ through a softmax over geometric scores (Eq.~\ref{eq:budget_softmax}), ensuring comprehensive global coverage across diverse semantic regions. 
(Right) \textbf{Stage-II (Micro-Mining)} computes sub-cluster sampling weights $W(S_j)$ by combining the inherited budget $r_k$, a semantic score modulated by a structural penalty $\exp(-\lambda \mathcal{L}_{struct}(S_j))$ (Eq.~\ref{eq:structural}), and a cohesion gate $[\beta_{S_j}+\epsilon]$ (Eq.~\ref{eq:geo_gate}). This stage refines selection within each region to downweight off-manifold outliers while preserving representative local structural dependencies.}
    \label{fig:framework}

\end{figure*}

\subsection{Problem Formulation and Overview}
\label{sec:problem_formulation}

Let $\mathcal{D}_{raw} = \{x_i\}_{i=1}^N$ denote the uncurated corpus. We employ an embedding model $f_\theta$ to map each sample $x_i$ to a normalized feature vector $\mathbf{e}_i \in \mathbb{S}^{d-1}$. We use this normalized space as a practical latent manifold $\mathcal{M}$ for clustering and selection.
We cast data selection not as distributional alignment to an external reference, but as a self-contained \textbf{manifold approximation} problem. Our objective is to identify a subset $S \subset \mathcal{D}_{raw}$ that maximally preserves the topological structure of $\mathcal{M}$ while suppressing off-manifold outliers.
Table \ref{tab:notation} summarizes the key notations utilized throughout this framework.
\begin{table}[h]
    \centering
    \small
    \renewcommand{\arraystretch}{1.1} 
    \caption{Summary of Key Notations and Symbols.}
    \label{tab:notation}
    \begin{tabularx}{\columnwidth}{p{1.4cm}X} 
        \toprule
        \textbf{Symbol} & \textbf{Definition} \\
        \midrule
        \rowcolor{gray!10} \multicolumn{2}{l}{\textbf{I. Macro-Exploration (Stage-I)}} \\
        $C_{k}$ & Global semantic cluster. \\
        $\mathbf{z}_k, s_k$ & Geometric features and score. \\
        $\mathbf{w}$ & Spectral consensus weights. \\
        $K^*$ & Optimal manifold resolution. \\
        $T_{scale}$ & Alignment scaling factor. \\
        $r_k$ & Global mixing budget. \\

        \rowcolor{gray!10} \multicolumn{2}{l}{\textbf{II. Micro-Mining (Stage-II)}} \\
        $S_{j}$ & Granular sub-cluster within $C_k$. \\
        $P_{S_j}$ & Semantic scores via probes. \\
        $\mathcal{L}_{\text{struct}}$ & Structural penalty. \\
        $\beta_{S_j}$ & Geometric cohesion gate. \\
        $W(S_j)$ & Final sampling weight. \\

        \rowcolor{gray!10} \multicolumn{2}{l}{\textbf{III. Manifold Theory \& Approximation}} \\
        $\mathcal{M}, d$ & Latent manifold and dimensionality. \\
        $\mathcal{E}(S)$ & Wasserstein-2 approximation error. \\
        $\sigma_k^2$ & Intra-cluster variance of $C_k$ in latent space. \\
        $\alpha_k$ & Cluster mass (probability) $\mu(C_k)$. \\
        $\Delta_{gain}^{(k)}$ & Variance mass rejected by Stage-II pruning. \\
        \bottomrule
    \end{tabularx}
\end{table}

\subsection{Stage-I: Macro Exploration (Global Clustering and Weighting)}
\label{sec:stage_1}

As illustrated in \textbf{Figure~\ref{fig:framework} (Left)}, Stage-I performs the \textbf{Macro-Distribution Balancing}. This phase transforms the raw manifold into structured semantic clusters and assigns a sampling budget to each cluster. The process consists of three sequential steps:

\paragraph{Geometric Metrics and Scoring.}
First, we characterize each candidate cluster $C_k$ via a feature vector $\mathbf{z}_k \in \mathbb{R}^4$. These dimensions act as geometric proxies:

\begin{itemize}[leftmargin=12pt, noitemsep]

    \item \textbf{Cohesion ($z_{\text{coh}}$):} Inverse intra-cluster distance. It prioritizes tight semantic structures akin to Neural Collapse~\citep{papyan2020prevalence} to ensure gradient stability.

    \item \textbf{Cluster Size ($z_{\text{size}}$):} Sample volume. Used to mitigate head redundancy (e.g., boilerplate) and shift focus to the information-dense long tail~\citep{huang2024demystifying}.

    \item \textbf{Sequence Length ($z_{\text{len}}$):} The average token count, a proxy for verbosity, used to downweight overly long, low-information clusters.

    \item \textbf{Entropy ($z_{\text{ent}}$):} Distributional impurity of language identifiers. It penalizes semantic ambiguity to ensure domain purity.

\end{itemize}

To unify these proxies, we combine the normalized signals with a weighted linear score to derive a scalar \textbf{Geometric Score} $s_k$. This score acts as a global quality metric, formulated as a contrast between structural coherence and statistical instability:
\begin{equation}
    \label{eq:geometric_score}
    s_k = w_{\text{coh}} \tilde{z}_{\text{coh}} - \sum_{f \in \mathcal{F}_{neg}} w_{f} \tilde{z}_{f}
\end{equation}
where $\mathcal{F}_{neg} = \{\text{len, ent, size}\}$ denotes the set of negative factors, and $\tilde{z}$ represents the Z-normalized magnitudes. The weighting vector $\mathbf{w}$ is derived via Spectral Consensus (see Appendix~\ref{sec:matrix}).

\paragraph{Intrinsic Resolution Selection.}
Central to this is a \textbf{Cross-Resolution Soft Alignment}. We construct a probabilistic bridge $\pi_{K \to K'}$ based on centroid similarity:
\begin{equation}
    \label{eq:transition_weight}
    \pi_{K \to K'}(j|i) = \text{Softmax}\left( T_{scale} \cdot \cos(\mu_i^{(K)}, \mu_j^{(K')}) \right)
\end{equation}
Using this bridge, we project the geometric scores $\mathbf{s}^{(K')}$ (derived in Eq.~\ref{eq:geometric_score}) from the finer resolution back to the current layer to obtain the \textbf{reconstructed scores} $\mathbf{\hat{s}}^{(K)}$:

\begin{equation}
    \label{eq:reconstruction}
    \hat{s}_i^{(K)} = \sum\nolimits_j \pi(j|i) s_j^{(K')}
\end{equation}
Stability is quantified via rank correlation using a  Kendall's $\tau$ (detailed formulation in Appendix~\ref{sec:K Selection}):
\begin{equation}
    \label{eq:stability_functional}
    J_{stab}(K) = \tau\left( \mathbf{s}^{(K)}, \, \mathbf{\hat{s}}^{(K)} \right)
\end{equation}
The resolution $K^*$ that maximizes this stability is selected as the intrinsic resolution (macro-granularity) via Algorithm~\ref{alg:intrinsic_resolution}.

\begin{algorithm}[t]
\caption{Intrinsic Resolution via Soft-Alignment}
\label{alg:intrinsic_resolution}
\begin{algorithmic}[1]
\REQUIRE Manifold $\mathcal{M}$, Range $\mathcal{K}$, Stride $\Delta K$.
\ENSURE Optimal resolution $K^*$.
\STATE $\mathcal{S} \leftarrow \emptyset$
\FOR{$K \in \mathcal{K}$}
    \STATE $K' \leftarrow K + \Delta K$
    \STATE \textbf{1. Extract:} Compute centroids $\mathcal{C}^{(K)}$ and scores $\mathbf{s}^{(K)}$.
    \STATE \textbf{2. Align:} Construct soft bridge $\pi_{K \to K'}$ (Eq.~\ref{eq:transition_weight}).
    \STATE \textbf{3. Evaluate:} Calculate rank stability $J_{stab}(K)$ (Eq.~\ref{eq:stability_functional}).
    \STATE $\mathcal{S} \leftarrow \mathcal{S} \cup \{ (K, J_{stab}(K)) \}$
\ENDFOR
\RETURN $K^* = \underset{K}{\arg\max} \{ J : (K, J) \in \mathcal{S} \}$
\end{algorithmic}
\end{algorithm}

\paragraph{Sampling Budget Allocation.}
With the optimal resolution $K^*$ established, we finally allocate the sampling budget $\mathbf{r}$. Leveraging the geometric scores from Eq.~(\ref{eq:geometric_score}), the probability for each cluster is computed via a standard softmax:
\begin{equation}
    \label{eq:budget_softmax}
    r_k = \frac{\exp(s_k)}{\sum_{j=1}^{K^*} \exp(s_j)}
    \end{equation}
This framework ensures the budget concentrates on semantically dense regions (high $s_k$).

\subsection{Stage-II: Micro-Mining (Local Sub-Clustering and Refinement)}
\label{sec:stage_2}

As illustrated in Figure~\ref{fig:framework} (Right), Stage-II performs the \textbf{Micro-Quality Selection} via local sub-clustering. This phase decomposes each global cluster $C_k$ into fine-grained sub-clusters $\{S_j\}$ to capture semantic diversity and topological details. The selection is refined through three coupled mechanisms: 

\paragraph{Probe-based Semantic Scoring.}

To assess semantic content efficiently, we extract a small representative \textbf{Probe Set} (indicated by the solid centroids in Figure~\ref{fig:framework}) from each sub-cluster $S_j$. We utilize an LLM as a \textbf{Knowledge Probe} to derive a \textbf{Semantic Score} $P_{S_j}$. Implementation details are provided in Appendix~\ref{sec:prompts}.

\paragraph{Relative Structural Consistency.}
To enforce manifold consistency, we define a \textbf{Structural Penalty} $\mathcal{L}_{struct}$ using a rectified Mahalanobis distance. It penalizes sub-clusters $S_j$ where the empirical mean $z_{f}(S_j)$ of features (e.g., length, entropy) exceeds the geometric consensus $(\mu_f, \sigma_f)$ of the parent cluster $C_k$:

\begin{equation}
    \label{eq:structural}
    \mathcal{L}_{struct} = \sum_{f \in \{len, ent\}} \left[ \frac{z_{f}^{(S_j)} - \mu_{f}^{(C_k)}}{\sigma_{f}^{(C_k)}} \right]_+^2
\end{equation}

We introduce $\mathcal{L}_{\text{struct}}$ not only to ensure that the curated subset preserves the underlying manifold topology but also to mitigate the score saturation observed in semantic scores (Fig.~\ref{fig:probe_quality} and Fig.~\ref{fig:subcluster_density}).

\paragraph{Geometric Cohesion Gate.}
To ensure topological robustness, we apply a \textbf{Geometric Cohesion Gate} $\beta_{S_j}$, depicted as the gating module in Figure~\ref{fig:framework}. Leveraging the cohesion metric $z_{coh}$ defined in Stage-I, we modulate sampling confidence based on the sub-cluster's compactness relative to its parent cluster:
\begin{equation}
    \label{eq:geo_gate}
    \beta_{S_j} = \text{Sigmoid}\left( z_{coh}^{(S_j)} - z_{coh}^{(C_k)} \right)
    \end{equation}
    
This gating mechanism suppresses sub-clusters with lower cohesion than their parent ($z_{coh}^{(S_j)} < z_{coh}^{(C_k)}$) while retaining structurally compact regions.


\paragraph{Hierarchical Sampling Weight.}
The final sampling probability $W(S_j)$ synthesizes the global budget $r_{C_k}$ with local metrics via a multiplicative modulation. It is computed as:
\begin{equation}
    \label{eq:final_weight}
    \resizebox{0.95\linewidth}{!}{%
    $W(S_j) \propto \underbrace{r_{k}}_{\text{Budget}} \cdot \underbrace{P_{S_j} \exp\left(-\lambda \mathcal{L}_{struct}(S_j)\right)}_{\text{Geometry-Aware Score}} \cdot \underbrace{[\beta_{S_j} + \epsilon]}_{\text{Gate}}$%
    }
\end{equation}

Within the geometry-aware score, we use a semantic score modulated by a geometric penalty to suppress off-manifold sub-clusters. Simultaneously, the gate $\left[ \beta_{S_j} + \epsilon \right]$ ensures cohesion, with $\epsilon$ preventing mode collapse by maintaining a minimal exploration floor.

\subsection{Theoretical Analysis: Manifold Approximation}
\label{sec:theory}

We frame data selection as minimizing the transport cost between the empirical measure $\hat{\mu}_S$ and the true manifold distribution $\mu$ \citep{villani2009optimal}. Let $f_\theta: \mathcal{X} \to \mathbb{R}^d$ be an $L$-Lipschitz embedding into a latent manifold of intrinsic dimension $d \ll D$ \citep{pope2021intrinsic, du2021learning}.

\noindent\textbf{Definition 1 (Approximation Error).}
We quantify the quality of subset $S$ by the squared Type-2 Wasserstein distance \citep{peyre2019computational}:
\begin{equation}
\label{eq:wasserstein_obj}
\begin{split}
    \mathcal{E}(S) &\triangleq W_2^2(\mu, \hat{\mu}_S) \\
    &= \inf_{\gamma \in \Pi(\mu, \hat{\mu}_S)} \iint_{\mathcal{M}^2} \|x - y\|^2 d\gamma(x, y).
\end{split}
\end{equation}
A constructive two-stage transport argument (Appendix~\ref{sec:proofs}) yields the following constant-factor decomposition:

\begin{equation}
\label{eq:decomposition}
\begin{aligned}
\mathcal{E}(S) \leq \;& 2 \underbrace{\sum_{k=1}^K \int_{C_k} \|x - \mathbf{c}_k\|^2 \, d\mu(x)}_{\text{Stage-I: Quantization}} \\
& + 2 \underbrace{\sum_{k=1}^K \alpha_k \mathbb{E}_{x \sim \hat{\mu}_{S_k}} \|x - \mathbf{c}_k\|^2}_{\text{Stage-II: Pruning}}.
\end{aligned}
\end{equation}

\noindent\textbf{Theorem 1 (UniGeM Bound).}
For density $p(x)$, the UniGeM approximation error satisfies:
\begin{equation}
\label{eq:theory_bound}
\mathcal{E}(S_{UniGeM}) \le 
2C_d K^{-2/d} + 2\sum_{k=1}^{K} \alpha_k\big(\sigma_k^2 - \Delta_{gain}^{(k)}\big),
\end{equation}
where $C_d$ is the Zador constant~\citep{zador1982asymptotic}, $\alpha_k \triangleq \mu(C_k)$, and $\sigma_k^2 \triangleq \mathbb{E}_{x \sim \mu(\cdot \mid C_k)} \|x - \mathbf{c}_k\|^2$.
Here $\Delta_{gain}^{(k)} \ge 0$ captures the within-cluster reduction induced by Stage-II pruning.
$\mathcal{L}_{struct}$ is a practical signal for identifying the outliers driving $\Delta_{gain}^{(k)}$. \noindent\textbf{Remark.}
The bound is derived under idealized modeling assumptions and is intended to guide the design rather than exactly characterize all engineering details.

\section{Experimental Setup}
\label{sec:exp_setup}


We evaluate UniGeM on \textbf{large-scale code pre-training}. Code is a demanding setting for data curation: small syntax issues can break execution \citep{li2023starcoder}, the corpus spans many programming languages \citep{feng2020codebert}, and programs follow hierarchical dependencies \citep{guo2020graphcodebert}. These properties make naive filtering brittle, and they let us test whether UniGeM preserves both coverage and local structure.

\subsection{Corpus Construction and Sampling Protocol}
\label{subsec:corpus}



We construct a 100B-token training corpus by mixing \textbf{The Stack Dedup} \citep{kocetkov2022stack} and Common Crawl \citep{Schaefer:2017:boilerplate-detection} in a fixed 7:3 code-to-text ratio. The 30B text component is frozen across all experiments.
This fixed ratio ensures that gains on code benchmarks stem from UniGeM's strategic data blending rather than a simple inflation of total code volume~\citep{xie2023doremi}.

\definecolor{epochblue}{RGB}{226,238,252}
\definecolor{highlightyellow}{RGB}{245,240,236}
\definecolor{upbg}{RGB}{220,245,220}
\definecolor{downbg}{RGB}{255,225,225}

\newcommand{\DeltaSup}[1]{%
  \edef\_d{\fpeval{round(#1,2)}}%
  \ifdim \_d pt = 0pt
  \else
    \ifdim \_d pt > 0pt
      \textsuperscript{\colorbox{upbg}{\textcolor{green!60!black}{\scriptsize$\uparrow$\num{\_d}}}}%
    \else
      \textsuperscript{\colorbox{downbg}{\textcolor{red!70!black}{\scriptsize$\downarrow$\num{\fpeval{abs(\_d)}}}}}%
    \fi
  \fi
}

\newcommand{\Score}[2]{%
  \num{#2}\DeltaSup{\fpeval{#2-(#1)}}%
}

\sisetup{
  round-mode=places,
  round-precision=1, 
  detect-weight=true,
  detect-family=true
}

\begin{table*}[t]
\centering
\small
\setlength{\tabcolsep}{1.8pt} 
\renewcommand{\arraystretch}{1.15}

\caption{Coding benchmark results. We compare the scaling properties between UniGeM-16B and UniGeM-8B under a constant inference budget (1.4B active parameters). The best scores in Block 2 and Block 3 are \underline{underlined}.}
\label{tab:coding_benchmarks}

\resizebox{\linewidth}{!}{%
    \begin{tabular}{ll|c|ccccccc}
    \toprule
    \multicolumn{2}{c}{\textbf{Setting}} &
    \textbf{Avg.} &
    \textbf{HE} & 
    \textbf{HE$^{+}$} &
    \textbf{MBPP} &
    \textbf{MBPP$^{+}$} &
    \textbf{LiveCode} &
    \multicolumn{2}{c}{\textbf{CruxEval}} \\
    \cmidrule(lr){1-2}\cmidrule(lr){3-3}\cmidrule(lr){4-8}\cmidrule(lr){9-10}
    \textbf{Methods} & \textbf{Epoch} &
    \textbf{Score} &
    \textbf{Pass@1} &
    \textbf{Pass@1} &
    \textbf{Pass@1} &
    \textbf{Pass@1} &
    \textbf{Pass@1} &
    \textbf{Input} & \textbf{Output} \\
    \midrule
    \addlinespace[0.5ex]
    
    \rowcolor{epochblue}
    \multicolumn{10}{c}{\textbf{UniGeM-16B (16B Total / 1.4B Active)}} \\
    \midrule
    \noalign{%
      \gdef\BOneAVG{32.92}
      \gdef\BOneHeA{55.9}%
      \gdef\BOneHeP{48.6}%
      \gdef\BOneMBPP{33.5}%
      \gdef\BOnePlus{37.6}%
      \gdef\BOneLCB{1.1}%
      \gdef\BOneCI{29.8}%
      \gdef\BOneCO{24.1}%
    }
    \multicolumn{1}{l}{Random} & 1.0 &
    \num[round-precision=1]{\BOneAVG} & \num{\BOneHeA} & \num{\BOneHeP} & \num{\BOneMBPP} & \num{\BOnePlus} & \num{\BOneLCB} & \num{\BOneCI} & \num{\BOneCO} \\
    \cmidrule(lr){1-10}
    \textbf{UniGeM (Ours)} & 0.5 &
    \Score{32.92}{32.31} & \Score{\BOneHeA}{49.9} & \Score{\BOneHeP}{43.8} & \Score{\BOneMBPP}{34.6} & \Score{\BOnePlus}{40.9} & \Score{\BOneLCB}{6.4} & \Score{\BOneCI}{26.0} & \Score{\BOneCO}{24.6} \\
    \rowcolor{highlightyellow}
    \textbf{UniGeM (Ours)} & 1.0 &
    \Score{32.92}{39.50} & \Score{\BOneHeA}{59.3} & \Score{\BOneHeP}{55.1} & \Score{\BOneMBPP}{41.8} & \Score{\BOnePlus}{47.8} & \Score{\BOneLCB}{10.8} & \Score{\BOneCI}{31.6} & \Score{\BOneCO}{30.1} \\
    \midrule
    
    \rowcolor{epochblue}
    \multicolumn{10}{c}{\textbf{UniGeM-8B (8B Total / 1.4B Active)}} \\
    \midrule
    \noalign{%
      \gdef\BTwoAVG{29.11}%
      \gdef\BTwoHeA{45.3}%
      \gdef\BTwoHeP{44.1}%
      \gdef\BTwoMBPP{30.5}%
      \gdef\BTwoPlus{34.2}%
      \gdef\BTwoLCB{2.0}%
      \gdef\BTwoCI{22.3}%
      \gdef\BTwoCO{25.5}%
    }
    \multicolumn{1}{l}{Random} & 1.0 &
    \num[round-precision=1]{\BTwoAVG} & \num{\BTwoHeA} & \num{\BTwoHeP} & \num{\BTwoMBPP} & \num{\BTwoPlus} & \num{\BTwoLCB} & \num{\BTwoCI} & \num{\BTwoCO} \\
    \cmidrule(lr){1-10}
    
    Meta-rater & 1.0 &
    \Score{\BTwoAVG}{35.02} & \Score{\BTwoHeA}{53.2} & \Score{\BTwoHeP}{49.6} & \Score{\BTwoMBPP}{35.3} & \Score{\BTwoPlus}{43.5} & \Score{\BTwoLCB}{4.7} & \underline{\Score{\BTwoCI}{31.4}} & \Score{\BTwoCO}{27.7} \\
    
    CLIMB & 1.0 &
    \Score{\BTwoAVG}{35.21} & \Score{\BTwoHeA}{52.4} & \Score{\BTwoHeP}{48.8} & \Score{\BTwoMBPP}{36.7} & \Score{\BTwoPlus}{45.5} & \Score{\BTwoLCB}{6.9} & \Score{\BTwoCI}{29.7} & \Score{\BTwoCO}{26.6} \\
    
    \textbf{UniGeM (Ours)} & 0.5 &
    \Score{\BTwoAVG}{30.02} & \Score{\BTwoHeA}{47.7} & \Score{\BTwoHeP}{43.5} & \Score{\BTwoMBPP}{32.4} & \Score{\BTwoPlus}{42.2} & \Score{\BTwoLCB}{5.2} & \Score{\BTwoCI}{16.6} & \Score{\BTwoCO}{22.5} \\
    
    \rowcolor{highlightyellow}
    \textbf{UniGeM (Ours)} & 1.0 &
    \underline{\Score{\BTwoAVG}{36.44}} & \underline{\Score{\BTwoHeA}{53.7}} & \underline{\Score{\BTwoHeP}{50.3}} & \underline{\Score{\BTwoMBPP}{37.4}} & \underline{\Score{\BTwoPlus}{46.6}} & \underline{\Score{\BTwoLCB}{7.8}} & \Score{\BTwoCI}{31.0} & \underline{\Score{\BTwoCO}{28.4}} \\
    \midrule
    
    \rowcolor{epochblue}
    \multicolumn{10}{c}{\textbf{Ablation Study (based on UniGeM-8B)}} \\
    \midrule
    \noalign{%
      \gdef\BThreeAVG{29.11}%
    }
    \multicolumn{1}{l}{Random} & 1.0 &
    \num[round-precision=1]{\BThreeAVG} & \num{45.3} & \num{44.1} & \num{30.5} & \num{34.2} & \num{2.0} & \num{22.3} & \num{25.5} \\
    \cmidrule(lr){1-10}
    
    Cluster Random & 1.0 &
    \Score{\BThreeAVG}{29.18} & \Score{45.3}{46.0} & \Score{44.1}{44.1} & \Score{30.5}{31.1} & \Score{34.2}{34.6} & \Score{2.0}{1.8} & \Score{22.3}{21.4} & \Score{25.5}{25.2} \\
    
    \textit{w/o} Hierarchy & 1.0 &
    \Score{\BThreeAVG}{33.00} & \Score{45.3}{51.4} & \Score{44.1}{48.1} & \Score{30.5}{34.5} & \Score{34.2}{40.1} & \Score{2.0}{1.9} & \Score{22.3}{27.0} & \Score{25.5}{28.0} \\
    
    \textit{w/o} Stage-II & 1.0 &
    \Score{\BThreeAVG}{31.21} & \Score{45.3}{45.8} & \Score{44.1}{43.8} & \Score{30.5}{33.8} & \Score{34.2}{40.1} & \Score{2.0}{2.3} & \Score{22.3}{24.3} & \Score{25.5}{28.3} \\
    
    \textit{w/o} Stage-I & 1.0 &
    \Score{\BThreeAVG}{34.86} & \Score{45.3}{52.2} & \Score{44.1}{49.5} & \Score{30.5}{36.1} & \Score{34.2}{43.1} & \Score{2.0}{4.5} & \underline{\Score{22.3}{31.3}} & \Score{25.5}{27.5} \\
    
    \rowcolor{highlightyellow}
    \textbf{UniGeM (Ours)} & 1.0 &
    \underline{\Score{\BThreeAVG}{36.44}} & \underline{\Score{45.3}{53.7}} & \underline{\Score{44.1}{50.3}} & \underline{\Score{30.5}{37.4}} & \underline{\Score{34.2}{46.6}} & \underline{\Score{2.0}{7.8}} & \Score{22.3}{31.0} & \underline{\Score{25.5}{28.4}} \\
    \bottomrule
    \end{tabular}%
}
\vspace{-10pt}
\end{table*}

\subsection{UniGeM Implementation Details}

Deploying geometric clustering on terabytes of data presents a computational challenge. We adopt a \textit{Probing-and-Scaling} strategy:

\begin{itemize}

\item \textbf{Phase I: Macro-Topology Discovery.} We first removed near-duplicates of evaluation benchmarks from the raw corpus. Subsequently, we embed the corpus using \textit{Qwen3-embedding} \citep{zhang2025qwen3}. To ensure scalability, we perform K-means clustering on a representative subset ($\mathcal{D}_{probe} \approx 20\%$) to identify the optimal macro-granularity $K^*=72$. Each sample in the full corpus is assigned to its nearest global centroid.

\item \textbf{Phase II: Hierarchical Geometric Mining.} Within each global cluster $C_k$, we execute localized sub-clustering to capture fine-grained micro-structures. The sub-cluster density $S_j$ is determined by the square root of the cluster population.
\end{itemize}

\subsection{Model and Hyperparameter}
\label{subsec:model}

We adopt a fine-grained sparse Mixture-of-Experts (MoE) Transformer, increasing the total expert capacity while keeping the number of activated parameters per token fixed to preserve inference efficiency~\citep{deepseek_v2}. We instantiate \textbf{UniGeM-8B} (32 experts) and \textbf{UniGeM-16B} (64 experts), both with \textbf{1.4B} active parameters. All models are trained \textbf{from scratch} with the same training recipe; only the data curation differs~\footnote{Reproducibility statement is discussed in Appendix \ref{app:reproduct}.}. For geometric selection, we use $\lambda=0.5$, $T_{scale}=20$, and $\epsilon=0.01$. See Appendix~\ref{sec:Experiment} for details.

\subsection{Compared Methods}
\paragraph{Baseline \& SOTA Methods}
\begin{itemize}[leftmargin=*, noitemsep, topsep=2pt]
    \item \textbf{Random Sampling:} Uniform sampling from the raw corpus as a reference baseline.
    \item \textbf{Meta-rater} \citep{zhuang2025metarater}: A representative \textit{instance-level selection} method relying on \textit{LLM-based scoring} to rank individual data quality.
    \item \textbf{Nemotron-CLIMB} \citep{diao2025climb}: A representative \textit{domain-level data mixing} method that optimizes global balancing weights.
\end{itemize}
To ensure competitive baselines, we apply a \textbf{code adaptation} to Meta-rater and CLIMB to reduce domain mismatch (Appendix~\ref{sec:Experiment}).

\paragraph{Ablation Variants}
To validate the hierarchical design, we compare UniGeM against:
\begin{itemize}[leftmargin=*, noitemsep, topsep=2pt]
    \item \textbf{Cluster Random:} Replaces the coarse-to-fine hierarchy with a single flat clustering stage ($K = K_{total}$) and performs random sampling with a uniform ratio in each cluster.
    \item \textbf{\textit{w/o} Hierarchy:} Uses the same flat clustering structure as above ($K = K_{total}$) but selects data via UniGeM's micro-mining metrics instead of random sampling.
    
    \item \textbf{\textit{w/o} Stage-I:} Assigns uniform global budgets ($r_k = 1/K$) while retaining Stage-II mining, utilizing \textbf{global feature statistics} as the reference for both the structural  and the geometric cohesion gate.
    
    \item \textbf{\textit{w/o} Stage-II:} Uses optimized global weights $\mathbf{r}$ but performs random sampling within clusters.
\end{itemize}

\begin{figure*}[t]
    \centering
    \makebox[\textwidth][c]{\includegraphics[width=1.1\linewidth]{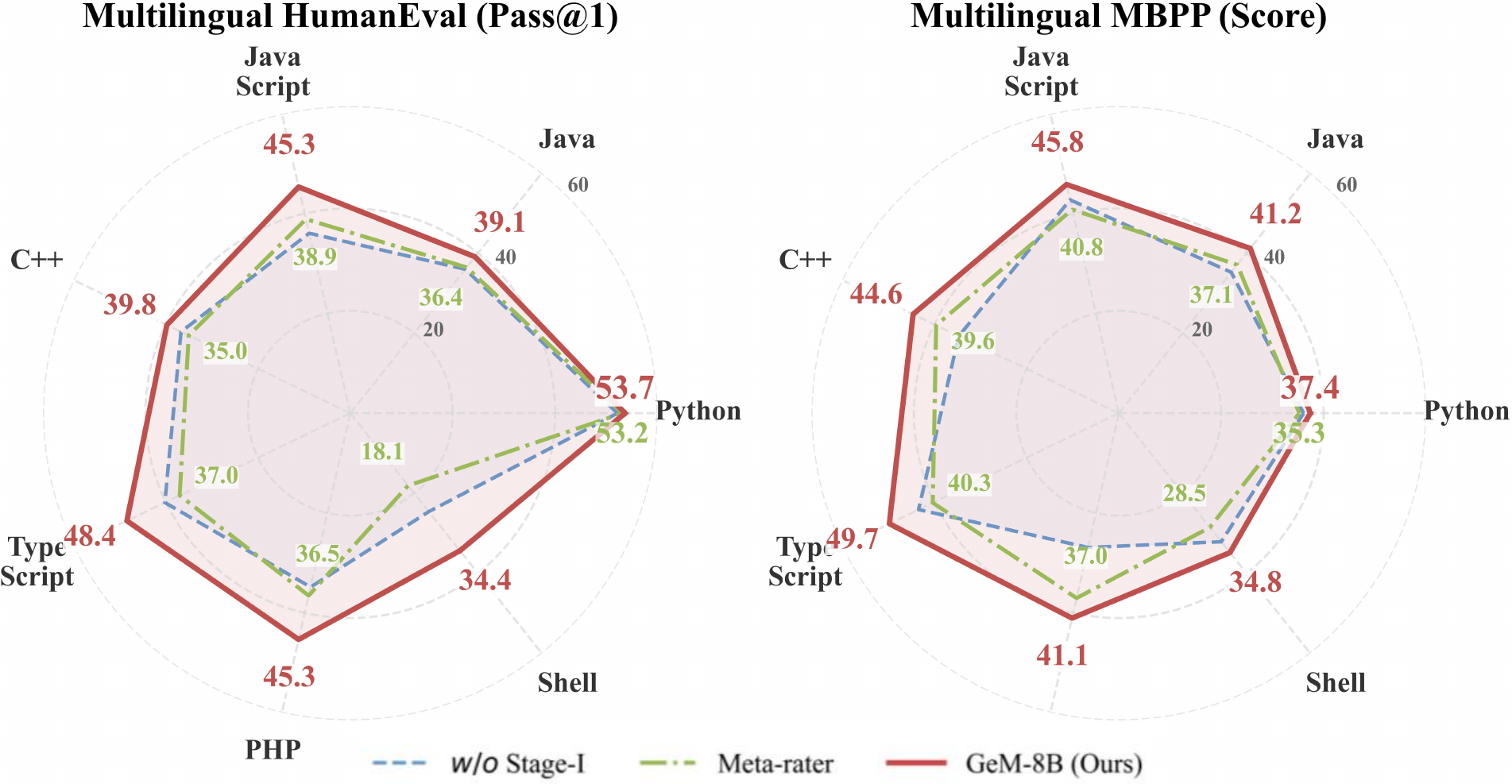}}    
    \caption{\textbf{Impact of Macro-Exploration (Stage-I).} UniGeM (red) outperforms Meta-rater (green) and the \textit{w/o Stage-I} variant (blue) across 7 languages, indicating that global clustering improves multilingual generalization.}
\label{fig:multilingual}
\end{figure*}

\subsection{Code Benchmarks}
We evaluate UniGeM on a diverse set of coding benchmarks covering code generation, robustUniGEMness/reasoning, and multilingual proficiency:
\begin{itemize}[leftmargin=*, noitemsep, topsep=2pt]
    \item \textbf{Code Generation:} We utilize \textit{HumanEval(+)} (\textbf{HE}) \citep{chen2021evaluating} and \textit{MBPP(+)} \citep{austin2021program} to evaluate generation performance.
    \item \textbf{Robustness \& Reasoning:} We assess temporal generalization using \textit{LiveCodeBench} (\textbf{LCB}) \citep{jain2024livecodebench} and execution prediction capabilities using \textit{CruxEval} \citep{gu2024cruxeval}.
    \item \textbf{Multilingual Proficiency:} We conduct a fine-grained analysis across multiple programming languages using \textit{MultiPL-E} \citep{cassano2023multiple}.
    
\end{itemize}


\section{Result Analysis}
\label{sec:result_analysis}


In this section, we evaluate UniGeM from multiple perspectives. We compare it with state-of-the-art baselines, examine cross-lingual generalization, and study the impact of the hierarchical design and training efficiency through ablation and scaling analyses.

\subsection{Comparison with State-of-the-Art}
\label{sec:sota}

We compare UniGeM against the Random baseline, \textbf{CLIMB} and \textbf{Meta-rater} on the 8B MoE model. Table~\ref{tab:coding_benchmarks} summarizes the results; the \textbf{Avg.} score represents the unweighted arithmetic mean of all reported metrics.

\paragraph{Overall Superiority and Data Efficiency.}
UniGeM achieves the best overall score (\textbf{36.4}), above the adapted domain-mixing baseline (\textbf{CLIMB}, 35.2) and the adapted instance-selection baseline (\textbf{Meta-rater}, 35.0). It is also more data-efficient: UniGeM reaches \textbf{30.0} after 0.5 epochs, surpassing Random sampling at 1.0 epoch (29.1), corresponding to an approximate 2.0$\times$ efficiency gain.
After 1.0 epoch, UniGeM further improves to \textbf{36.4}, indicating that the gain is not limited to early training.

\paragraph{Reasoning and Generalization.}
UniGeM's advantage is particularly robust on complex execution and out-of-distribution (OOD) tasks. On \textbf{LiveCodeBench}, UniGeM scores \textbf{7.8}, clearly surpassing CLIMB (6.9) and significantly exceeding Meta-rater (4.7). 
On \textbf{CruxEval}, the adapted Meta-rater remains highly competitive and slightly leads on \textbf{Input} prediction, suggesting that code-aware LLM scoring captures common logical patterns. UniGeM performs best on \textbf{Output} execution prediction (\textbf{CruxEval-Output}) with \textbf{28.4}, indicating better coverage of more complex execution behaviors.

\subsection{Multilingual Proficiency}
\label{sec:multilingual}

We evaluate cross-lingual generalization on \textbf{Multilingual HumanEval} and \textbf{MBPP} (Fig.~\ref{fig:multilingual}). UniGeM consistently dominates the Meta-rater and \textit{w/o Stage-I} variant across 7 languages. On HumanEval, UniGeM achieves decisive leads in low-resource domains like \textbf{Shell} (\textbf{34.4\%} vs. Meta's 18.1\%) and \textbf{TypeScript} (\textbf{48.4\%} vs. 37.0\%), while maintaining clear advantages in strict-syntax languages like \textbf{C++} (\textbf{39.8\%} vs. 35.0\%) and \textbf{Java} (\textbf{39.1\%} vs. 36.4\%). This superiority persists on MBPP (e.g., \textbf{44.6\%} vs. 39.6\% in C++), confirming UniGeM's robust, language-agnostic transfer capabilities.

\subsection{Ablation and Hyperparameter Analysis}
\label{sec:ablation}

We conduct ablations and sensitivity analyses on UniGeM-8B to isolate the contribution of each component.

\paragraph{Contribution of Model Components.}
Ablations highlight the role of each stage. Removing Stage-I (\textit{w/o Stage-I}) replaces learned global budgets with uniform allocation, which weakens global coverage; performance stays relatively stable on high-resource languages (e.g., Python) but drops on under-represented languages such as Shell and PHP in multilingual evaluation. Conversely, removing Stage-II (\textit{w/o Stage-II}) reduces HumanEval Pass@1 from 53.7\% to 45.8\%, indicating that local geometric refinement is important even with optimized macro mixing.

\paragraph{Statistical Properties of Stage-I Features.}
Statistical analysis confirms that extensive features (e.g., length) follow a Log-Normal distribution while intensive features (e.g., cohesion) are naturally stable, supporting our hybrid geometric priors (details in Appendix~\ref{sec:features_statistics}).

\paragraph{Sensitivity to Global Granularity ($K$).}

\begin{figure}[t]
    \centering
    \includegraphics[width=1.0\linewidth]{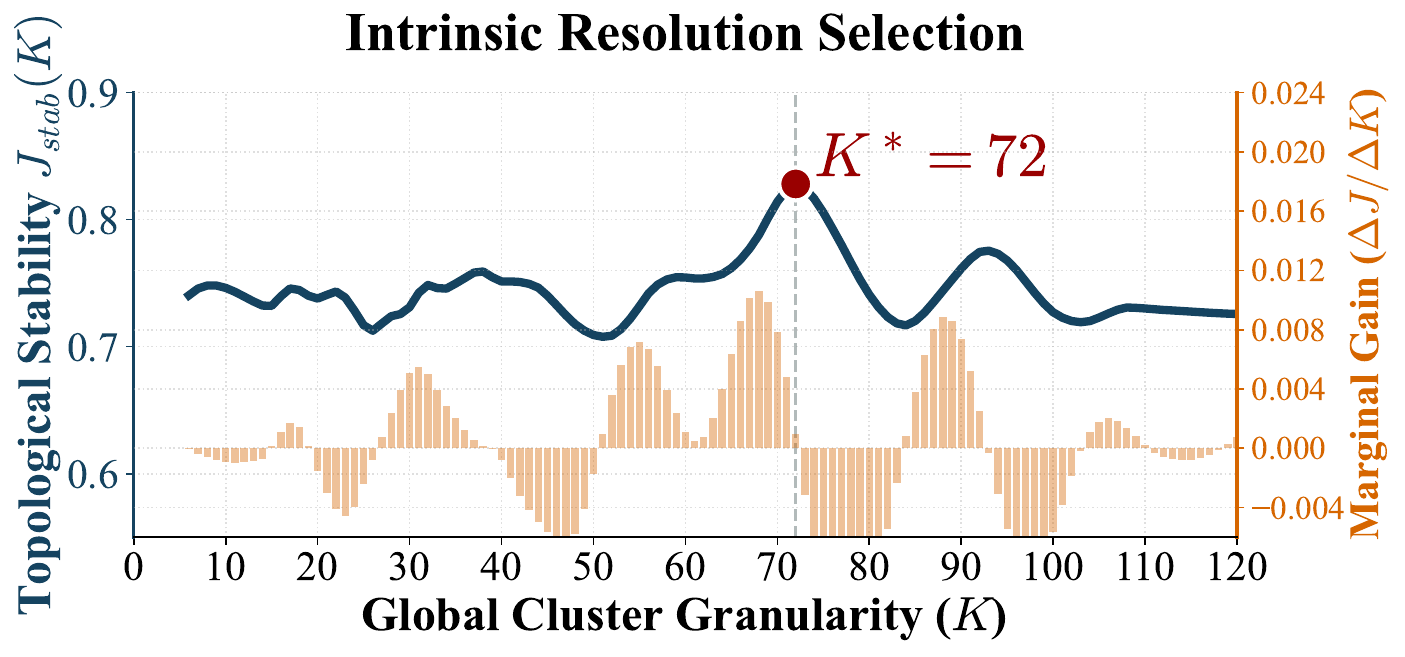}
    \caption{\textbf{Intrinsic Resolution Selection.} The stability index $J_{\text{stab}}(K)$ (blue line) rises and then plateaus, attaining its maximum at $K^*=72$. The marginal gain (orange bars) diminishes significantly beyond this point.}
    \label{fig:k_selection}
\end{figure}

The cluster number $K$ controls the resolution of the global approximation. As shown in Figure~\ref{fig:k_selection}, the stability index $J_{stab}(K)$ increases and then plateaus beyond $K \approx 60$. We choose $K^*=72$ as a stable setting with finer semantic resolution; smaller $K$ (e.g., $K<40$) suffers from under-segmentation, merging distinct semantic domains and reducing stability. Beyond $K^*$, the marginal gain ($\Delta J$) diminishes to near zero, suggesting that 72 clusters are sufficient to capture the corpus's semantic structure without over-segmentation or fragmentation.

\begin{figure}[t]
    \centering
    \includegraphics[width=1.0\linewidth]{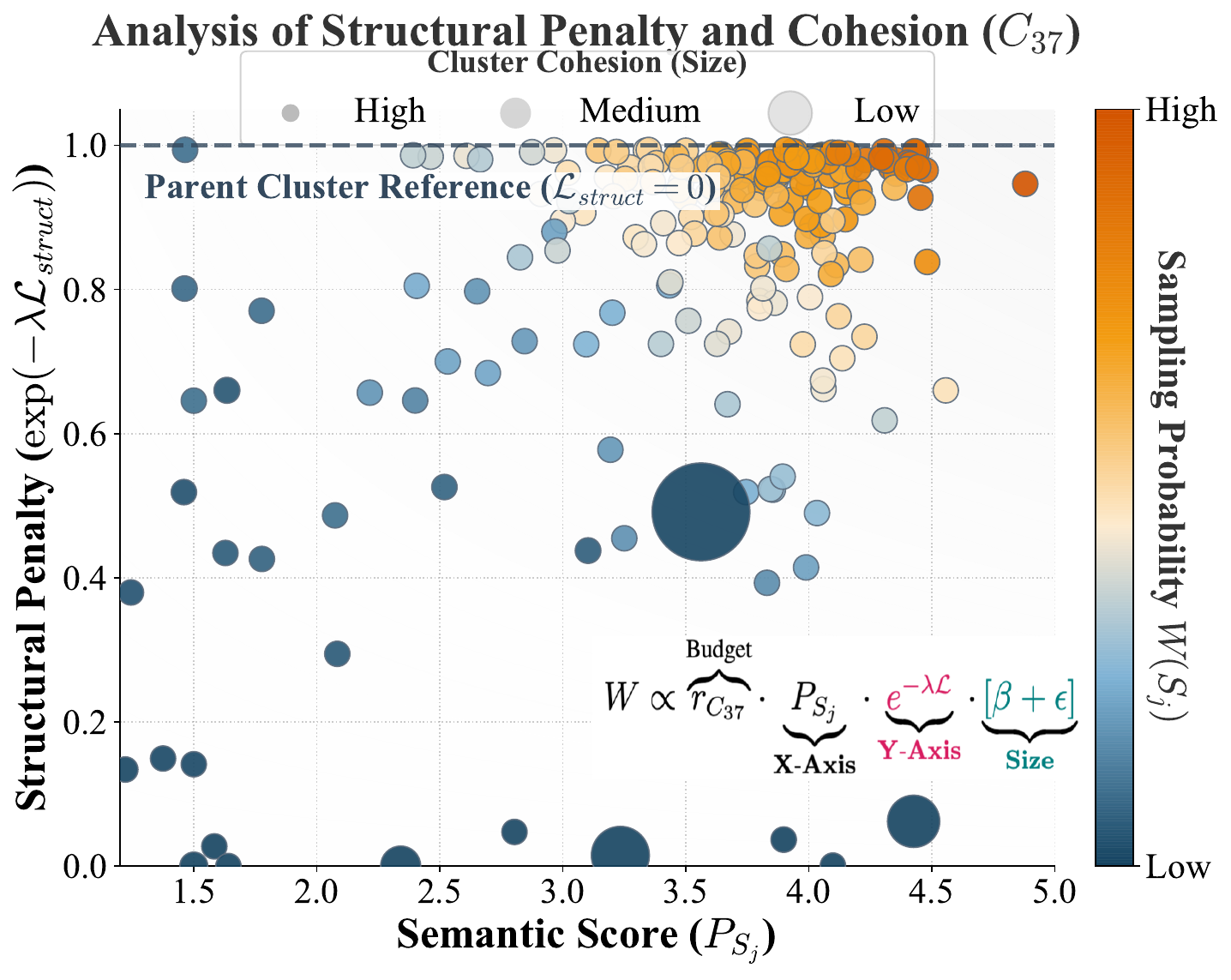}
    \caption{Empirical distribution of a representative subset of sub-clusters in $C_{37}$. Sub-clusters are mapped by \textbf{Semantic Score} ($P_{S_j}$, X-axis) and \textbf{Structural Penalty} (Y-axis), with bubble size representing \textbf{Cluster Cohesion}. The color gradient indicates the \textbf{Sampling Probability} $W(S_j)$, illustrating the budget allocation toward sub-clusters.}
    
    \label{fig:structural_Penalty}
\end{figure}

\paragraph{Analysis of Structural Penalty and Cohesion.} 

Figure~\ref{fig:structural_Penalty} illustrates how UniGeM balances semantic utility and structural consistency at the sub-cluster level. Sub-clusters with higher \textbf{Semantic Score} and a small structural deviation (i.e., \textbf{Structural Penalty} close to 1) receive higher sampling probability, whereas sub-clusters that score well semantically but deviate structurally are downweighted. \textbf{Cluster Cohesion} (bubble size) further biases sampling toward compact, well-formed sub-clusters that better represent the parent cluster.

\subsection{Scaling Analysis}
\label{sec:scaling}

\paragraph{Model Scaling Dynamics.}

We further study scaling by training a larger \textbf{UniGeM-16B} model. At 0.5 epochs, UniGeM-16B reaches \textbf{32.3}, which is lower than its 1.0-epoch score, consistent with larger models being more data-hungry~\citep{hoffmann2022chinchilla}. After 1.0 epoch, UniGeM-16B improves to \textbf{39.5} (\textbf{+6.6} over Random), showing that the curated corpus continues to benefit training as model size increases.

\section{Related Work}

Current work on data efficiency is moving beyond static filtering toward methods that model dataset structure. One prominent line trains \textbf{proxy models} to derive mixing or importance weights (e.g., DoReMi~\citep{xie2023doremi} and DCLM~\citep{li2024scalingfilter}), but this can be compute-heavy and may introduce ``proxy bias,'' where signals from small proxies do not transfer cleanly to larger target models~\citep{mindermann2022prioritized,sorscher2022beyond}. A related family relies on reference datasets for alignment or selection~\citep{xie2023dsir,mass}, while model-aware approaches such as Mates~\citep{yu2024mates,zhang2025harnessing} use influence-style estimates to capture sample-level contributions. To better preserve reasoning ability, QuaDMix~\citep{liu2025quadmix} explicitly balances quality and diversity, though gradient- or training-intensive signals can limit scalability~\citep{mindermann2022prioritized,li2024scalingfilter}. More principled directions aim to avoid expensive training signals altogether: DDOQ~\citep{tan2025dataset} casts selection as pushforward optimal quantization, improving over heuristic clustering schemes~\citep{chen2023skill,diao2025climb}, and Wasserstein-manifold views model dataset dynamics in a way that goes beyond flat domain mixing toward preserving structure relevant for complex reasoning~\citep{atanackovic2024meta}.

\section{Conclusion}
\label{sec:Conclusion}

We introduced \textbf{UniGeM}, a hierarchical framework that unifies macro-distribution balancing and micro-quality selection through \textit{manifold approximation}. By using topological stability to choose the global resolution and geometric priors for instance mining, UniGeM curates a compact, structure-preserving training set from code corpora. Experiments with 8B and 16B MoE models show \textbf{2.0$\times$ data efficiency} over a random baseline and \textbf{better one-epoch performance} than strong adapted baselines, with consistent gains in code reasoning and multilingual evaluations.


\section*{Limitations}
\label{sec:Limitation}
Despite its effectiveness, this work has several limitations: \begin{enumerate}[leftmargin=*, noitemsep, topsep=0pt] 

\item \textbf{Domain Specificity:} Our evaluation focused primarily on the \textbf{code corpus}. While code provides a rigorous testbed for geometric structures, the efficacy of UniGeM on massive, heterogeneous \textbf{general web text} mixtures remains to be fully explored. 

\item \textbf{Computational Overhead:} The initial global embedding and clustering phase, while mitigated by our probing-and-scaling strategy, still requires non-trivial resources when applied to trillion-token scales.

\item \textbf{Static Pipeline:} The current framework operates as a pre-processing step. Future work is required to integrate UniGeM into online training pipelines to allow for dynamic, cross-domain manifold updates as the model's data needs evolve. 

\end{enumerate}

\bibliography{custom}

\appendix
\section{Unsupervised Hyperparameter Derivation}
\label{sec:matrix}

\begin{figure}[t]
    \centering
    \includegraphics[width=1\linewidth]{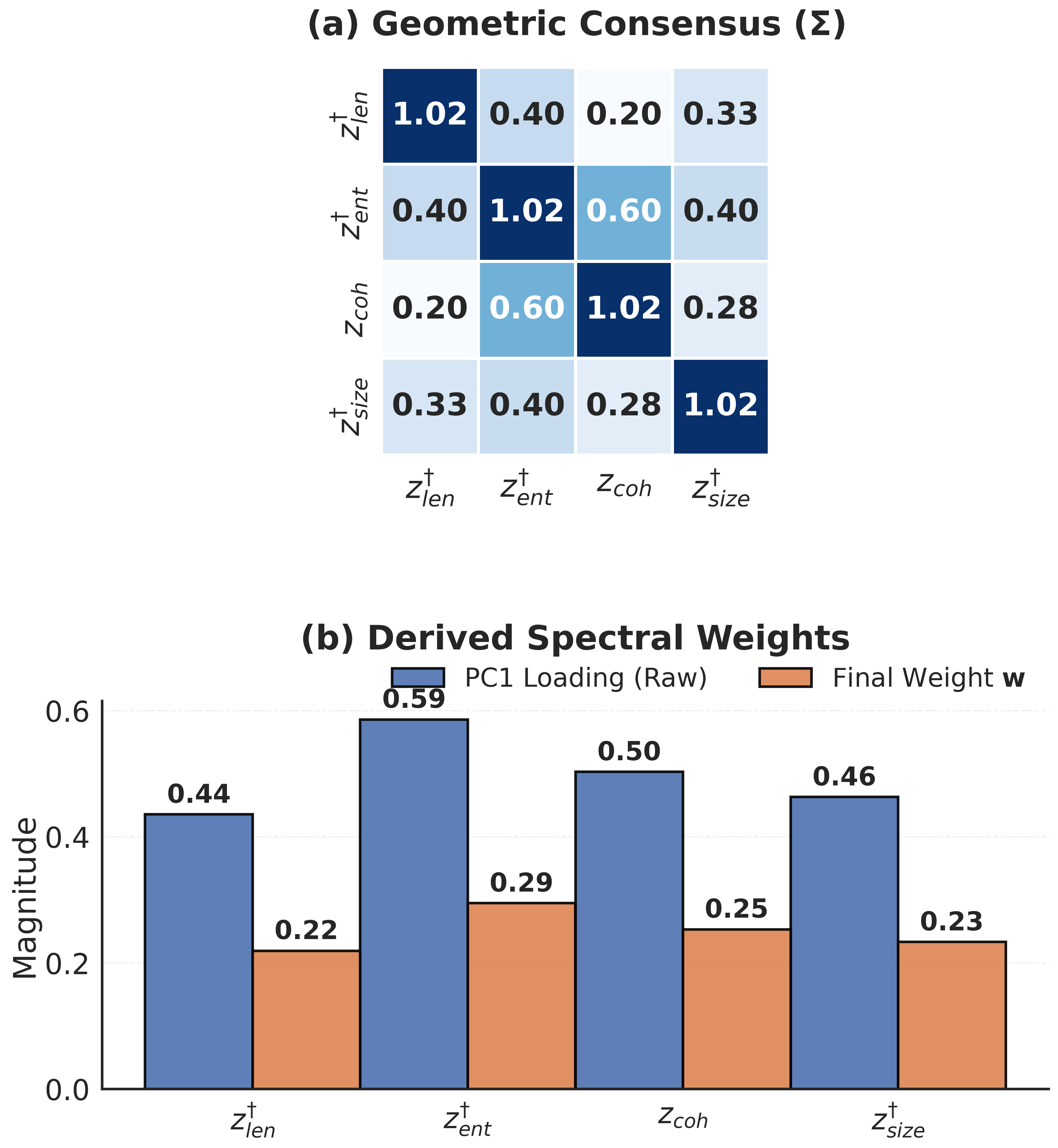}
    \caption{\textbf{Unsupervised Spectral Weight Derivation.} 
    \textbf{(a)} The geometric consensus matrix $\mathbf{\Sigma}$, computed on the \textbf{polarity-aligned metrics} (where penalties are inverted), reveals strong positive correlations (e.g., $0.60$ between aligned entropy and cohesion). This suggests that these proxies share a common latent direction.
    \textbf{(b)} Instead of heuristic tuning, we derive the balancing coefficients $\mathbf{w}$ directly from the first principal component (PC1). The spectral analysis intrinsically yields data-driven weights in which \textbf{Entropy} and \textbf{Cohesion} receive the largest coefficients.}
    
    \label{fig:spectral_analysis}
\end{figure}

\begin{figure}[t]
    \centering
    \includegraphics[width=1\linewidth]{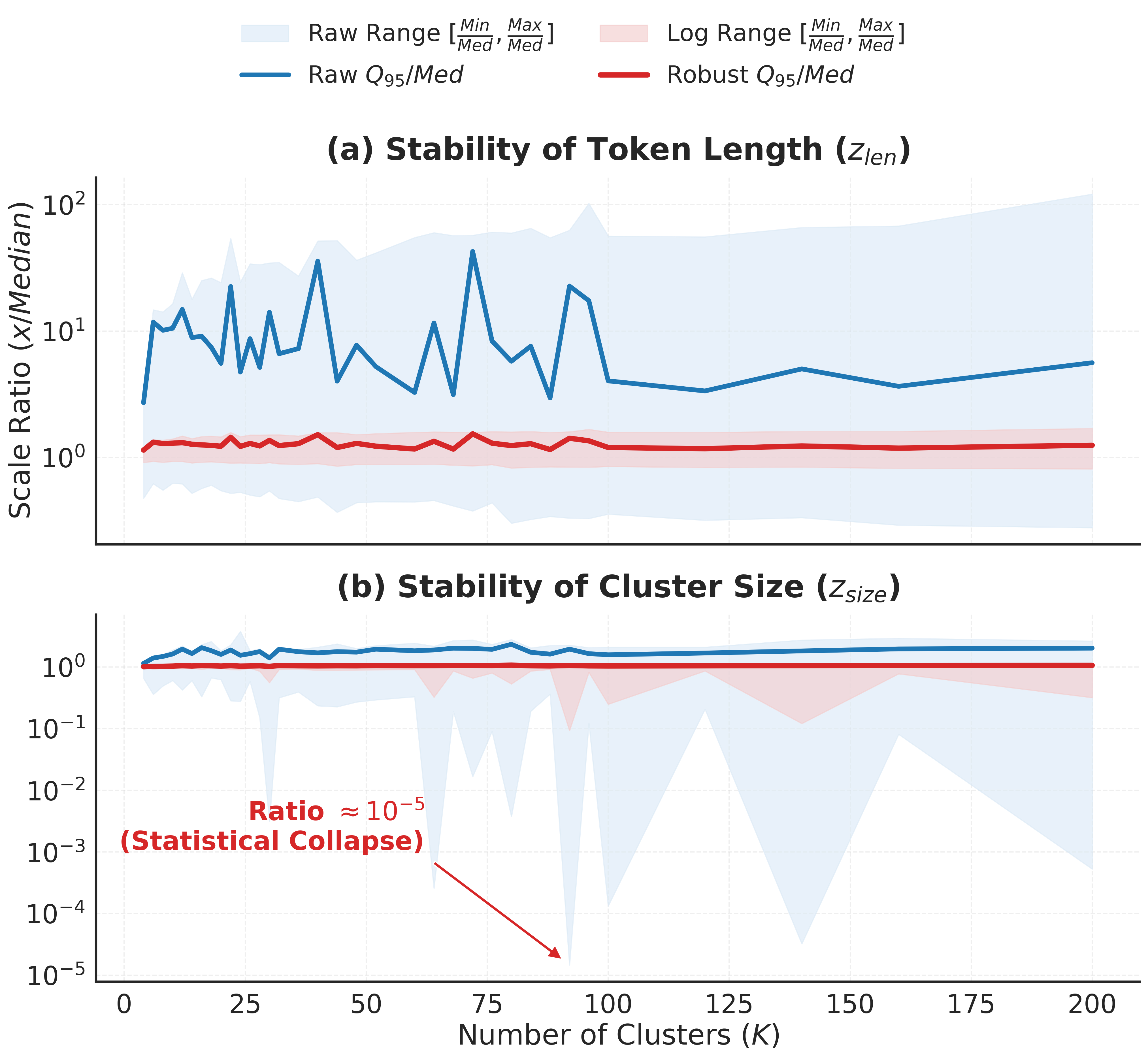}
    \caption{\textbf{Statistical Stability and the Necessity of Log Transform.} 
    We analyze the stability of feature statistics across varying cluster resolutions ($K \in [5, 200]$). 
    \textbf{(a)} For token length, the raw scale (blue) exhibits high-frequency oscillations, while our \textbf{log-stabilized estimator} ($z_{len}^\dagger$, red) remains smooth.
    \textbf{(b)} For cluster size, the raw statistics suffer from extreme \textbf{statistical collapse} (highlighted by the arrow, where the \textbf{minimum-to-median ratio} drops to $\sim 10^{-5}$), which would cause numerical instability in standard Z-score calculations.}
    \label{fig:robustness_check}
\end{figure}

In Eq.~(\ref{eq:geometric_score}), the scoring function relies on a weight vector $\mathbf{w}$ to balance diverse geometric features. Instead of relying on heuristic grid search, we derive $\mathbf{w}$ intrinsically from the data geometry. This process is grounded in the statistical stability analysis and spectral consensus visualized in Figure~\ref{fig:robustness_check} and Figure~\ref{fig:spectral_analysis}.

\paragraph{1. Feature Stabilization (Log-Normal Priors).}
Let $\mathcal{F} = \{z_{\text{coh}}, z_{\text{len}}, z_{\text{ent}}, z_{\text{size}}\}$ be the set of raw feature vectors. As visualized in \textbf{Figure~\ref{fig:robustness_check}}, raw extensive metrics (Length, Size) exhibit heavy-tailed instabilities. To mitigate this, we first apply log transform to stabilize their magnitudes.
We define the \textbf{stabilized feature matrix} $\mathbf{Z}^{\dagger}$ as follows:
\begin{equation}
\label{eq:dagger_def}
\begin{aligned}
    z_{\text{coh}}^{\dagger} &= z_{\text{coh}} \\
    z_{\text{ent}}^{\dagger} &= z_{\text{ent}} \\
    z_{\text{len}}^{\dagger} &= \log(z_{\text{len}}) \\
    z_{\text{size}}^{\dagger} &= \log(z_{\text{size}})
\end{aligned}
\end{equation}
Note that at this stage, $z^{\dagger}$ represents the raw physical properties (e.g., larger $z_{\text{len}}^{\dagger}$ still means longer sequence).

\paragraph{2. Standardization and Main Text Notation.}
We then apply Z-score standardization to the entire matrix $\mathbf{Z}^{\dagger}$ to ensure all dimensions share a unified scale (zero mean, unit variance). 
This yields the normalized metrics $\mathbf{\tilde{z}}$ utilized in the main text (Eq.~\ref{eq:geometric_score}):
\begin{equation}
    \tilde{z}_{k, f} = \text{Z-Score}(z_{k, f}^{\dagger})
\end{equation}
Thus, $\tilde{z}$ preserves the original polarity of the features. This is why Eq.~\ref{eq:geometric_score} explicitly subtracts the penalty terms (Length, Entropy, Size) to convert them into a quality score.

\paragraph{3. Spectral Weight Derivation.}
To derive the consensus weights $\mathbf{w}$, we construct a temporary \textbf{Aligned Matrix} $\mathbf{X}_{align}$ where all features are oriented towards "quality" (flipping the signs of  terms):
\begin{equation}
    \mathbf{X}_{align} = [\tilde{z}_{\text{coh}}, -\tilde{z}_{\text{ent}}, -\tilde{z}_{\text{len}}, -\tilde{z}_{\text{size}}]
\end{equation}
We then compute the covariance matrix $\mathbf{\Sigma} = \frac{1}{K-1} \mathbf{X}_{align}^\top \mathbf{X}_{align}$. As shown in \textbf{Figure~\ref{fig:spectral_analysis}(a)}, this alignment reveals strong positive correlations. The first principal eigenvector $\mathbf{v}_1$ of $\mathbf{\Sigma}$ captures the direction of \textbf{Maximum Consensus}. The final weights are derived by $L_1$-normalizing this vector: $\mathbf{w} = \mathbf{v}_1 / \|\mathbf{v}_1\|_1$.

\section{Intrinsic Resolution Selection Details}
\label{sec:K Selection}

\begin{figure*}[t]
    \centering
    \includegraphics[width=1\linewidth]{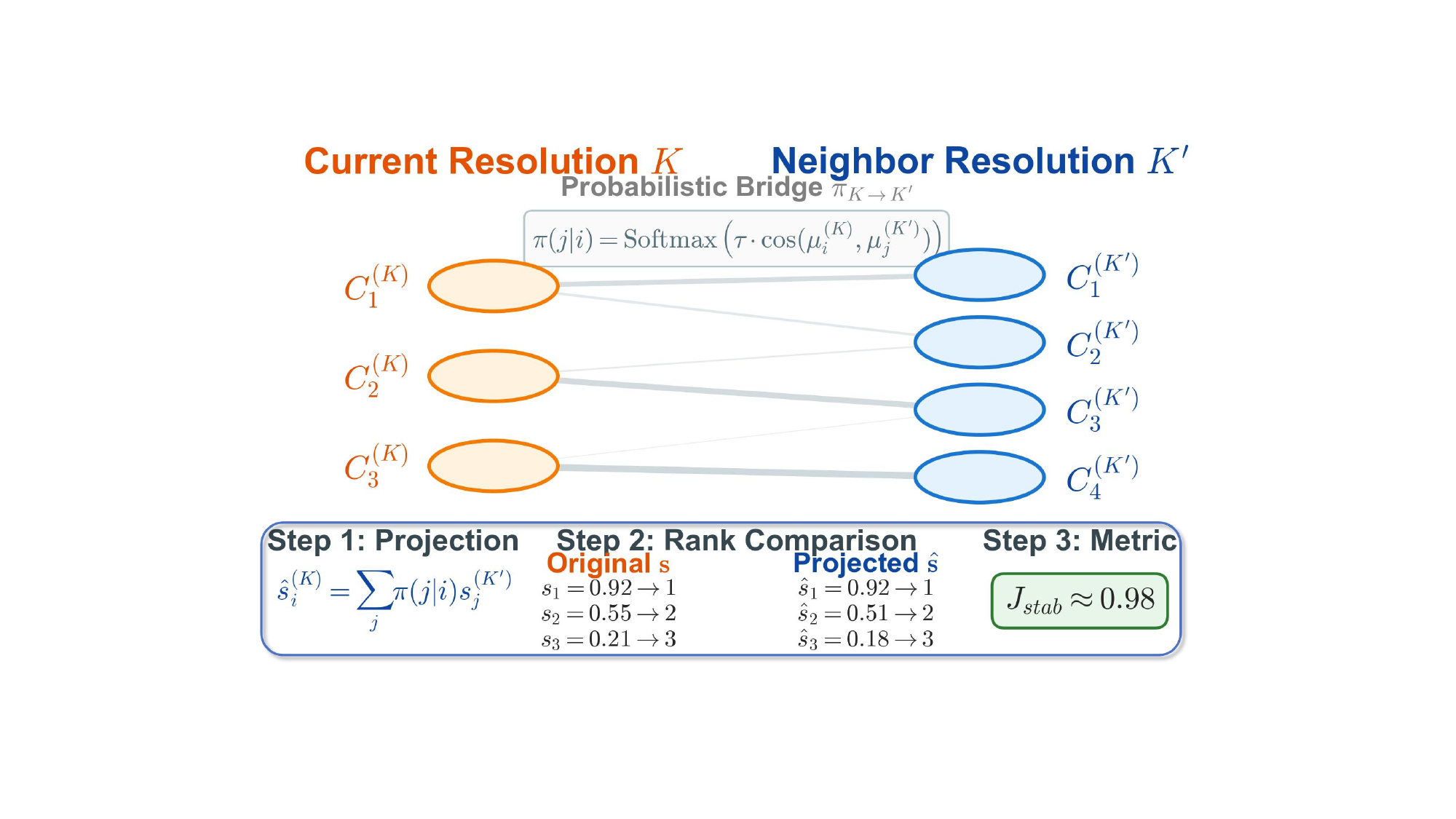}
    \caption{\textbf{Schematic of Intrinsic Resolution Selection.} 
    The workflow proceeds in three stages corresponding to the mathematical derivation: 
    (1) \textbf{Projection:} A probabilistic bridge $\pi$ is constructed via centroid similarity (Eq.~\ref{eq:transition_weight}) to back-project scores from neighbor resolution $K'$, yielding the reconstructed profile $\hat{s}^{(K)}$ (Eq.~\ref{eq:reconstruction}); 
    (2) \textbf{Rank Comparison:} The relative ordering of the original geometric scores $\mathbf{s}^{(K)}$ is compared against the reconstructed proxies $\mathbf{\hat{s}}^{(K)}$; 
    (3) \textbf{Metric Computation:} The final stability score $J_{stab}$ is derived using Kendall's $\tau$ rank correlation (Eq.~\ref{eq:stability_functional}). Numerical values (e.g., 0.92) are schematic examples for illustration.}
    \label{fig:schematic_appendix}
\end{figure*}

\subsection{Metric Definition: Rank Stability}
To robustly quantify the topological stability $J_{stab}(K)$ (rank stability across resolutions), we utilize Kendall's Rank Correlation Coefficient ($\tau$). This metric evaluates whether the relative quality ranking of clusters remains consistent after projecting to a finer resolution.

Given the two scoring vectors $\mathbf{s}^{(K)}$ and $\mathbf{\hat{s}}^{(K)}$ (from Eq.~\ref{eq:reconstruction}), we examine all possible pairs of clusters $(i, j)$ where $1 \le i < j \le K$. A pair is classified based on the consistency of their relative ordering:
\begin{align}
    \text{Concordant} &: (s_i - s_j)(\hat{s}_i - \hat{s}_j) > 0 \\
    \text{Discordant} &: (s_i - s_j)(\hat{s}_i - \hat{s}_j) < 0
\end{align}
Let $N_{conc}$ and $N_{disc}$ denote the total counts of such pairs:

\begin{align*}
    N_{\text{conc}} &= \sum_{i<j} \mathbb{I}(\text{Concordant}) \\
    N_{\text{disc}} &= \sum_{i<j} \mathbb{I}(\text{Discordant})
\end{align*}

where $\mathbb{I}(\cdot)$ is the indicator function. The final stability metric is the normalized difference between these counts:
\begin{equation}
    \label{eq:stability_functional_appendix}
    J_{stab}(K) = \frac{N_{conc} - N_{disc}}{K(K-1)/2} = \frac{2(N_{conc} - N_{disc})}{K(K-1)}
\end{equation}
The denominator $K(K-1)/2$ represents the total number of unique pairs. Thus, $J_{stab} \approx 1$ indicates that the geometric structure is perfectly preserved across resolutions.

\subsection{Implementation Optimizations}
To further ensure stability against sample size variations, we implement the following engineering optimizations:

\paragraph{1. Multi-Scale Hop Averaging.}
Instead of relying solely on the immediate neighbor ($K \to K+1$), which may be noisy, we compute stability across multiple strides $\Delta K \in \{2, 4, 6\}$. The final stability is a weighted average:
\begin{equation}
    J_{final}(K) = \frac{\sum_{h \in \Delta K} \gamma_h \cdot J_{stab}(K \to K+h)}{\sum \gamma_h}
\end{equation}
where $\gamma_d$ are decay weights (e.g., $[0.5, 0.3, 0.2]$) that prioritize local consistency.

\paragraph{2. Small-Sample Fisher Shrinkage.}
When $K$ is small, rank-correlation estimates can have high variance and appear overly optimistic. We therefore apply an \textbf{atanh-based shrinkage heuristic} to damp inflated stability scores in low-$K$ regimes.
We treat $J_{stab}\in(-1,1)$ as a bounded rank-stability score and map it with $z=\operatorname{arctanh}(J_{stab})$ for shrinkage; this is an engineering correction rather than a statistical guarantee.
\begin{equation}
    z_{shrunk} = z \cdot \tanh\left( \lambda_{shrink} \cdot \sqrt{N_{valid}-3} \right),
\end{equation}

\begin{equation} 
r_{\text{shrunk}} = \tanh(z_{\text{shrunk}}).
\end{equation}
where $N_{valid}$ denotes the effective number of clusters participating in the rank comparison and $\lambda_{shrink}$ is a regularization parameter. 

\section{Theoretical Proofs}
\label{sec:proofs}

In this section, we provide the detailed derivation for Theorem~1. We model the data selection process as an optimal quantization problem on a Riemannian manifold, drawing connections to recent theoretical advances in data pruning \citep{sorscher2022beyond}. All bounds below hold for an arbitrary number of clusters $K$; in UniGeM we instantiate $K=K^*$, where $K^*$ is selected by maximizing the stability objective $J_{stab}(K)$ (Algorithm~\ref{alg:intrinsic_resolution}).

\subsection{Proof of Error Decomposition}

Let $Q: \mathcal{M} \to \{\mathbf{c}_1, \dots, \mathbf{c}_K\}$ be the quantization operator mapping any point $x$ to its nearest global centroid $\mathbf{c}_{k(x)}$. Let $C_k$ denote the Voronoi cell induced by $\mathbf{c}_k$, and define $\alpha_k \triangleq \mu(C_k)$.

Starting from the definition of the squared Wasserstein-2 distance \citep{peyre2019computational},
\begin{equation}
\begin{aligned}
    \mathcal{E}(S) &= W_2^2(\mu, \hat{\mu}_S) \\
    &= \inf_{\gamma \in \Pi(\mu, \hat{\mu}_S)} \int_{\mathcal{M}\times\mathcal{M}} \|x-y\|^2 \, d\gamma(x,y)
\end{aligned}
\end{equation}
we upper bound $\mathcal{E}(S)$ by constructing an explicit two-stage transport plan through a centroid-supported intermediate measure.
Define $\mu_K \triangleq \sum_{k=1}^K \alpha_k \, \delta_{\mathbf{c}_k}$. By the triangle inequality of $W_2$,
\begin{equation}
    W_2(\mu, \hat{\mu}_S) \le W_2(\mu, \mu_K) + W_2(\mu_K, \hat{\mu}_S),
\end{equation}
and using $(a+b)^2 \le 2a^2 + 2b^2$ yields the constant-factor bound
\begin{equation}
    \label{eq:decomp_const_factor}
    \mathcal{E}(S) \le 2 W_2^2(\mu, \mu_K) + 2 W_2^2(\mu_K, \hat{\mu}_S).
\end{equation}

\noindent\textbf{Stage-I (Global distortion).}
Couple each $x\sim \mu$ with its centroid $Q(x)$ to obtain
\begin{equation}
    W_2^2(\mu, \mu_K)
    \le \sum_{k=1}^K \int_{C_k} \|x-\mathbf{c}_k\|^2 \, d\mu(x)
    \triangleq \mathcal{T}_1.
\end{equation}

\noindent\textbf{Stage-II (Within-cluster residual).}
For analysis, we view the selected-set empirical measure as a cluster-wise mixture
\begin{equation}
\hat{\mu}_S \triangleq \sum_{k=1}^K \alpha_k \, \hat{\mu}_{S_k},
\end{equation}
where $\hat{\mu}_{S_k}$ is the empirical measure supported on $S_k$.
Let $S_k \triangleq S \cap C_k$ be the selected subset inside cluster $C_k$. The second term $W_2^2(\mu_K,\hat{\mu}_S)$ measures how well the selected points within each $C_k$ represent the local mass anchored at $\mathbf{c}_k$. This induces a within-cluster residual term that we summarize by

\begin{equation}
W_2^2(\mu_K, \hat{\mu}_S)
\le
\sum_{k=1}^K \alpha_k \, \mathbb{V}(S_k)
\triangleq \mathcal{T}_2.
\end{equation}

where we define the within-cluster \emph{residual energy} (a second moment w.r.t. the centroid) as
\begin{equation}
    \mathbb{V}(S_k)\triangleq \mathbb{E}_{x\sim \hat{\mu}_{S_k}}\|x-\mathbf{c}_k\|^2.
\end{equation}

Combining the two parts with Eq.~(\ref{eq:decomp_const_factor}) yields a constructive decomposition of $\mathcal{E}(S)$ into a global quantization term $\mathcal{T}_1$ and a local within-cluster term $\mathcal{T}_2$, up to constant factors commonly used in quantization-style analyses \citep{gray2002quantization}.

\noindent\textbf{Remark (High-dimensional intuition).}
In high-dimensional embeddings ($d \gg 1$), cross-terms between centroid error and within-cluster residual are often empirically small due to concentration effects \citep{ledoux2001concentration}, which motivates the near-additive behavior observed in practice; however, our bound above does not rely on this approximation.

\subsection{Bound Derivation for Stage-I ($\mathcal{T}_1$)}

The first term $\mathcal{T}_1$ corresponds to the classical high-resolution quantization error. According to \textbf{Zador's Theorem} \citep{zador1982asymptotic}, for a quantizer with $K$ codepoints on a $d$-dimensional manifold with probability density function $p(x)$, the asymptotic distortion satisfies:
\begin{equation}
    \lim_{K \to \infty} K^{2/d} \cdot \mathcal{T}_1 = J_{d} \|p\|_{d/(d+2)},
\end{equation}
where $J_d$ is the coefficient of the optimal lattice quantizer in $\mathbb{R}^d$. This yields the Stage-I term in our bound:
\begin{equation}
    \mathcal{T}_1 \le C_d \cdot K^{-2/d}.
\end{equation}

\textbf{Implication:} This suggests that Stage-I controls global covering distortion through the choice of $K$.

\subsection{Bound Derivation for Stage-II ($\mathcal{T}_2$)}

The second term $\mathcal{T}_2$ represents the intra-cluster residual variance. For a standard random sampler, this corresponds to the raw cluster variance, which we denote by $\sigma_k^2$ for cluster $C_k$. Let $p_k(x) \triangleq p(x)/\alpha_k$ for $x \in C_k$ denote the conditional density within cluster $C_k$, where $\alpha_k=\mu(C_k)$.

\noindent\textbf{Assumption (dominant structural filtering).}
Within each cluster $C_k$, we assume the auxiliary reweighting terms used in practice (e.g., probe score $P_{S_j}$ and cohesion gate $\beta_{S_j}$) are either (i) approximately independent of the radial deviation $\|x-\mathbf{c}_k\|$ or (ii) bounded and do not systematically favor higher-deviation points.
Under this assumption, the dominant geometric effect of Stage-II is governed by $\exp(-\lambda \mathcal{L}_{struct}(x))$.

UniGeM modulates the sampling probability via $P(x) \propto \exp(-\lambda \mathcal{L}_{struct}(x))$. For theoretical analysis, we approximate this soft exponential decay as a truncation mechanism on an effective acceptance region
\begin{equation}
    \Omega_{UniGeM} = \{x \in C_k \mid \mathcal{L}_{struct}(x) < \tau \},
\end{equation}
where $\tau$ is a confidence threshold implicitly controlled by $\lambda$.

Define the random-baseline within-cluster second moment as
\begin{equation}
    \sigma_k^2 \triangleq \int_{C_k} \|x - \mathbf{c}_k\|^2 p_k(x)\, dx.
\end{equation}
Under the truncation approximation, UniGeM induces the conditional density
$q_k(x) \triangleq \frac{p_k(x)\mathbf{1}[x\in \Omega_{UniGeM}]}{Z_k}$ with
$Z_k \triangleq \int_{\Omega_{UniGeM}} p_k(x)\,dx$, and the resulting second moment is
\begin{equation}
    \mathbb{V}(S_{UniGeM}^{(k)}) \approx \int_{\Omega_{UniGeM}} \|x - \mathbf{c}_k\|^2 q_k(x)\, dx.
\end{equation}
We define the pruning gain as
\begin{equation}
    \Delta_{gain}^{(k)} \triangleq \sigma_k^2 - \int_{\Omega_{UniGeM}} \|x - \mathbf{c}_k\|^2 q_k(x)\, dx \;\;\ge 0,
\end{equation}
where the non-negativity holds when the acceptance region preferentially keeps lower-deviation points.

\subsection{Final Theorem Assembly and Remark}

Substituting the bounds for $\mathcal{T}_1$ and $\mathcal{T}_2$ back into the decomposition yields:

\begin{equation}
    \mathcal{E}(S_{UniGeM}) \leq 2C_d K^{-2/d} + 2\sum_{k=1}^K \alpha_k \big(\sigma_k^2 - \Delta_{gain}^{(k)}\big).
\end{equation}

\noindent\textbf{Practical proxy.}
In Eq.~(\ref{eq:theory_bound}), the gain $\Delta_{gain}^{(k)}$ captures the variance mass removed by pruning within $C_k$.
In practice, high-deviation samples correspond to large standardized feature deviations; thus $\mathcal{L}_{struct}(x)$ (a rectified squared Mahalanobis distance) directly serves as a proxy for identifying the rejected region $\Omega^c$ and monitoring pruning strength.

\section{Annotation Model.}
\label{sec:prompts}
We leverage the Qwen3-235B model as a \textbf{Knowledge Probe} to inspect the semantic attributes of these samples. Adopting a \textbf{Model-Based Annotation} strategy~\citep{seed2025seed}, we design a structured system prompt to decompose the analysis into four complementary dimensions: \textit{Code Quality}, \textit{Engineering Design}, \textit{Training Suitability}, and \textit{Knowledge Density}. This multi-dimensional rubric aims to capture both syntactic correctness and training-relevant content.

The specific system prompt used for this annotation is provided below.
\begin{promptbox}[System Prompt: Code Data Evaluation Strategy]
You are an expert code evaluator. Your task is to assess the provided code snippet based on four distinct dimensions. For each dimension, assign a score on a scale of 1 to 5 according to the criteria below.

\textbf{1. Code Quality \& Compliance} (Focus: Syntax, naming, and readability)
\begin{itemize}[leftmargin=12pt, noitemsep, topsep=0pt, partopsep=0pt]
    \item \textbf{1-2 (Low):} Severe syntax errors or logical flaws; chaotic naming conventions; inconsistent indentation; contains dead code or empty functions; non-executable.
    \item \textbf{3-4 (Mid):} Syntactically correct and executable; reasonable naming and formatting; minor redundancies or loose structure but generally readable.
    \item \textbf{5 (High):} Error-free; strict adherence to standards (e.g., PEP8); precise naming; concise logic with no redundancy; highly readable and linear structure.
\end{itemize}

\textbf{2. Algorithmic \& Engineering Design} (Focus: Modularity, robustness, and system thinking)
\begin{itemize}[leftmargin=12pt, noitemsep, topsep=0pt, partopsep=0pt]
    \item \textbf{1-2 (Low):} Monolithic structure (global scope); lack of modularity; absence of input validation or exception handling; hardcoded constants; fails on edge cases.
    \item \textbf{3-4 (Mid):} Basic modularity (function splitting); separation of concerns; foundational error checking; clear structure but lacks advanced abstraction or extensibility.
    \item \textbf{5 (High):} High-level abstraction (classes/patterns); robust engineering (exception safety, resource management); extensible, testable, and demonstrates system-level thinking.
\end{itemize}

\textbf{3. Training Suitability} (Focus: Educational value and style consistency for LLMs)
\begin{itemize}[leftmargin=12pt, noitemsep, topsep=0pt, partopsep=0pt]
    \item \textbf{1-2 (Low):} Arbitrary naming; missing or misleading comments; mixed styles; hallucinated or fragmentary code; lacks context.
    \item \textbf{3-4 (Mid):} Normative naming; comments cover key steps; consistent style; readable logic suitable for beginners or general training.
    \item \textbf{5 (High):} Strict adherence to community best practices; insightful comments explaining design intent; exemplary structure suitable for teaching or high-quality fine-tuning.
\end{itemize}

\textbf{4. Knowledge Density} (Focus: Technical insights and domain expertise)
\begin{itemize}[leftmargin=12pt, noitemsep, topsep=0pt, partopsep=0pt]
    \item \textbf{1-2 (Low):} Trivial operations (e.g., simple I/O, loops); lacks domain knowledge; low information density equivalent to introductory tutorials.
    \item \textbf{3-4 (Mid):} Standard algorithms or patterns (e.g., DFS, HashMaps); implements common best practices; explicit but generic knowledge.
    \item \textbf{5 (High):} Non-trivial insights or cross-domain knowledge (e.g., memory alignment, bitwise optimization, lock-free structures); reveals underlying principles or deep optimizations.
\end{itemize}

\textbf{Output Format:}
Output strictly in valid JSON format:
\texttt{\{"code\_quality": int, "algorithm\_and\_engineering": int, "training\_suitability": int, "knowledge\_score": int\}}

\textbf{Data Input:}
\$content
\end{promptbox}

\section{Experiment Details}
\label{sec:Experiment}

In this appendix, we provide the condensed technical specifications for our experiments, including the model architecture, training recipe, and adaptation protocols for SOTA baselines.

\subsection{Model Architectures}
We employ a fine-grained sparse MoE architecture with identical expert parameterization across models. We scale total capacity by increasing the number of experts, while keeping the routing strategy fixed (Top-2) so that the per-token activated parameter budget remains constant, preserving inference throughput.

\begin{table}[h]
    \centering
    \small
    \renewcommand{\arraystretch}{1.1}
    \begin{tabularx}{\columnwidth}{lXX}
        \toprule
        \textbf{Configuration} & \textbf{UniGeM-8B} & \textbf{UniGeM-16B} \\
        \midrule
        Total / Active Params & 8.0B / 1.4B & 16.8B / 1.4B \\
        Total Experts ($N$) & 32 & 64 \\
        Routing Strategy & Top-2 & Top-2 \\
        Hidden / Layers & 2048 / 24 & 2048 / 24 \\
        \bottomrule
    \end{tabularx}
\end{table}

\subsection{Pre-training Configuration}
The models are trained from scratch on a 100B-token mixture, consisting of 70B code tokens and 30B code-related text tokens. We utilize \textbf{Qwen3-235B} to retrieve the text component from Common Crawl to ensure semantic relevance. The training employs a Warmup-Stable-Decay (WSD) schedule.

\paragraph{Hyperparameter Selection Logic.}
The geometric hyperparameters $\{\lambda, T_{scale}, \epsilon\}$ are calibrated based on the statistical moments of the 20\% probe manifold. We set the \textbf{structural } $\lambda=0.5$ to ensure that samples deviating beyond $2\sigma$ from the geometric consensus incur a significant weight reduction (e.g. reduced to $\approx 13.5\%$ of the original weight), effectively pruning logical outliers. The \textbf{scale factor} $T_{scale}=20$ is employed to amplify the density contrast during cross-resolution alignment, ensuring the stability-driven clustering captures sharp manifold boundaries. Finally, an \textbf{exploration floor} $\epsilon=0.01$ is maintained to preserve long-tail distributional diversity and mitigate manifold approximation errors.

\begin{table}[h]
    \centering
    \small
    \renewcommand{\arraystretch}{1.1}
    \begin{tabularx}{\columnwidth}{lX}
        \toprule
        \textbf{Hyper-parameter} & \textbf{Value} \\
        \midrule
        Optimizer & AdamW ($\beta_1=0.9, \beta_2=0.95$, WD=0.1) \\
        Peak Learning Rate & $3.0 \times 10^{-4}$ (8B) / $2.4 \times 10^{-4}$ (16B) \\
        Batch Size / Seq Len & 2,560 $\to$ 8,960 / 4,096 Tokens \\
        Stability Mechanisms & NormHead, Stochastic Routing Warmup \\
        \midrule
        \rowcolor{gray!10} \multicolumn{2}{l}{\textbf{UniGeM Geometric Mining Params}} \\
        Structural ($\lambda$) & 0.5 (calibrated to $2\sigma$ consensus) \\
                
        Exploration Floor ($\epsilon$) & 0.01 (1\% diversity reserve) \\
        
        Transition Scale ($T_{scale}$) & 20 (Eq. \ref{eq:transition_weight}) \\
        
        \rowcolor{gray!10} \multicolumn{2}{l}{\textbf{Clustering \& Embedding}} \\
        Embedding Model & \textit{Qwen3-embedding} \\
        Global Resolution ($K^*$) & 72 (via stability-driven selection) \\
        K-means Iterations & 10 (Stage-I) / 5 (Stage-II) \\
        \bottomrule
    \end{tabularx}
\end{table}

\subsection{SOTA Baseline Adaptation}

Standard implementations of CLIMB and Meta-rater are designed for general text and often rely on proxy tasks (e.g., MMLU) or generic encoders (e.g., BERT), which can introduce domain mismatch for code. Using these methods strictly off-the-shelf would make the comparison less informative. We therefore apply a \textbf{Code Adaptation} protocol to improve their alignment with code and obtain strong code-aware baselines.

\begin{itemize}[leftmargin=*, noitemsep, topsep=0pt]
    \item \textbf{Nemotron-CLIMB Adaptation:} We replace general BERT embeddings with \textbf{Qwen3-embedding} to perform code-aware semantic clustering ($K=100$). The optimization target is shifted from MMLU to a \textbf{Code Oracle ($V_{\text{code}}$)}, which computes weighted validation loss on \textbf{MBPP-Sanitized}, \textbf{HumanEval-Pack}, and logic-dense samples from \textbf{DS-1000}.
    
    \item \textbf{Meta-rater Adaptation:} We re-define the PRRC framework into \textbf{Code-PRRC}, focusing on \textit{Professionalism} (complexity), \textit{Readability} (style), \textit{Reasoning} (flow density), and \textit{Cleanliness} (syntax). We score 500k seed samples via \textbf{Qwen3-235B} to distill four specialized \textbf{ModernBERT-base} scorers capable of handling long-context code quality assessment.
    
    \item \textbf{Common Proxy Setup:} To search for optimal weights, both methods utilize a \textbf{350M Dense Transformer} as a proxy model. These proxies are trained on 2B token slices across multiple iterations (64 trials for CLIMB; 256 for Meta-rater) to fit a LightGBM-based quality-to-loss regressor.
\end{itemize}

\section{Distributional Analysis of Features for Stage-I Cluster}
\label{sec:features_statistics}

In this section, we provide a detailed distributional analysis of the geometric proxies observed in our large-scale pre-training experiment. Specifically, we visualize the feature statistics across the optimal resolution of \textbf{$K^*=72$ global clusters}, identified via the Topological Stability analysis in Section~\ref{sec:ablation}. These empirical results validate the statistical assumptions underpinning our Geometric Scoring function (Eq.~\ref{eq:geometric_score}).

\begin{figure*}[h]
    \centering
    \includegraphics[width=1.0\linewidth]{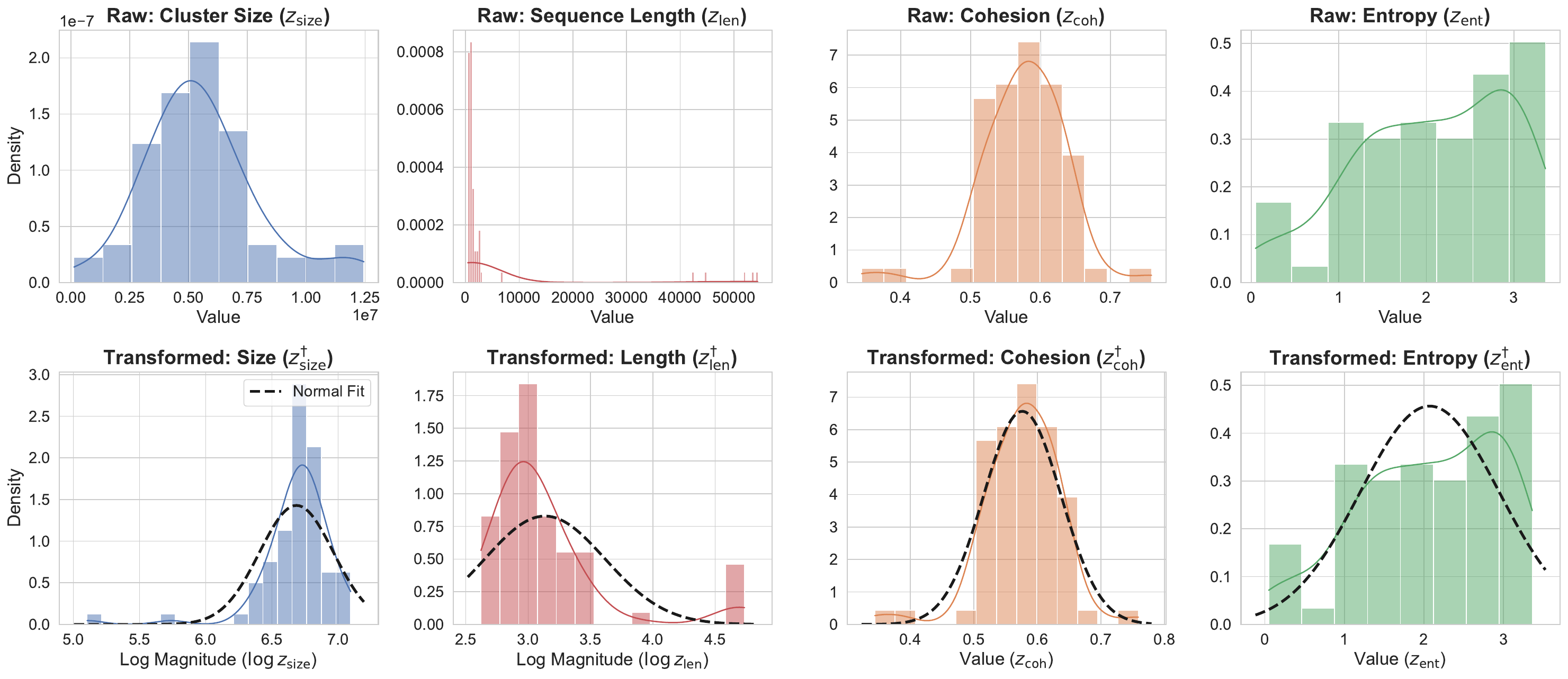}
    \caption{\textbf{Distributional Transformation across the $K^*=72$ Global Clusters.} 
    \textbf{(Top Row)} The raw distributions of extensive properties—\textit{Cluster Size} ($z_{\text{size}}$) and \textit{Sequence Length} ($z_{\text{len}}$)—exhibit extreme heavy-tailed skewness across the 72 experimental clusters.
    \textbf{(Bottom Row)} Applying the logarithmic transformation stabilizes these features; their log-values are closer to a Gaussian shape (as illustrated by the overlaid normal fits).
    In contrast, the intensive properties \textit{Cohesion} and \textit{Entropy} naturally follow a stable unimodal distribution, confirming the robustness of the extracted latent manifolds.}
    \label{fig:feature_distributions}
\end{figure*}

\paragraph{Transformation of Extensive Properties.}
As illustrated in the top row of Figure~\ref{fig:feature_distributions}, the raw distributions of \textbf{Cluster Size} ($z_{\text{size}}$) and \textbf{Sequence Length} ($z_{\text{len}}$) exhibit significant right-skewness across the 72 latent domains. This confirms our hypothesis in Section~\ref{sec:stage_1} that the raw code corpus is highly heterogeneous, spanning multiple orders of magnitude. Direct usage of these raw metrics would result in \textit{variance dominance}, where spectral analysis is biased by magnitude outliers rather than structural quality.
However, as shown in the bottom row, applying the logarithmic transformation effectively projects these features into log-space, where the transformed values are closer to Gaussian. This supports the normality assumption behind Z-score standardization and the subsequent spectral consensus step.

\paragraph{Stability of Intensive Properties.}
Conversely, the intensive properties—\textbf{Cohesion} ($z_{\text{coh}}$) and \textbf{Entropy} ($z_{\text{ent}}$)—naturally display bounded, unimodal distributions across the 72 clusters without transformation. This implies that the experimentally identified clusters are statistically well-formed, maintaining consistent internal densities and semantic purities. Consequently, we validate our hybrid processing strategy: while extensive features require logarithmic dampening to mitigate scale disparities, intensive features can be directly utilized as linear geometric priors to preserve their original sensitivity.

\section{Distributional Analysis of Semantic Scoring}
\label{sec:probe_statistics}

In this section, we report the statistical characteristics of the semantic scores obtained during the Stage-II exploration. We visualize both the raw discrete ratings from the Annotation Model and the aggregated continuous scores for sub-clusters to provide a comprehensive view of the data quality distribution.

\subsection{Probe and Sub-Cluster Score Distributions}
\begin{figure*}[t]
    \centering
    \includegraphics[width=1.05\linewidth]{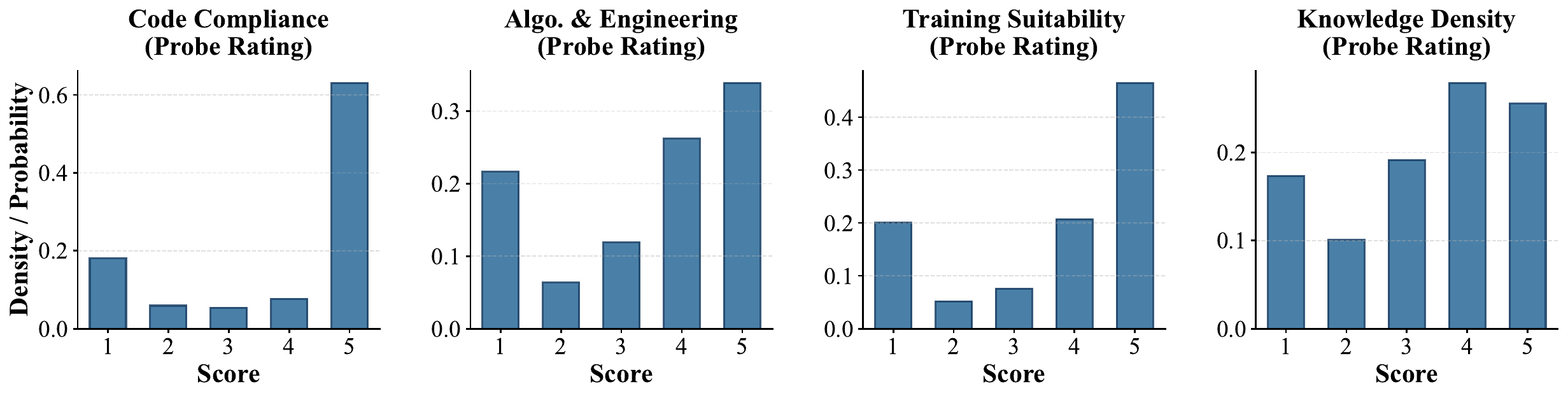}
    \vspace{-0.2cm}
    \caption{\textbf{Discrete Probe Quality Distribution (Raw Prompt Outputs).} 
    The probability mass distribution of integer scores (1--5) assigned by the \textbf{Annotation Model} to probe samples. The data exhibits a strong skew towards the upper bound (scores 4 and 5) across all dimensions, reflecting the general high acceptance rate of the judge model for code syntax.}
    \label{fig:probe_quality}
\end{figure*}

Figure~\ref{fig:probe_quality} presents the distribution of discrete quality ratings (scale 1--5) assigned by the Annotation Model to individual probe samples. The results indicate a pronounced left-skewed distribution across all four evaluation dimensions. Specifically, the \textit{Code Quality} and \textit{Training Suitability} metrics show a high concentration of samples receiving perfect or near-perfect scores ($>60\%$ rated as 5).

\begin{figure*}[t]
    \centering
    \includegraphics[width=1.05\linewidth]{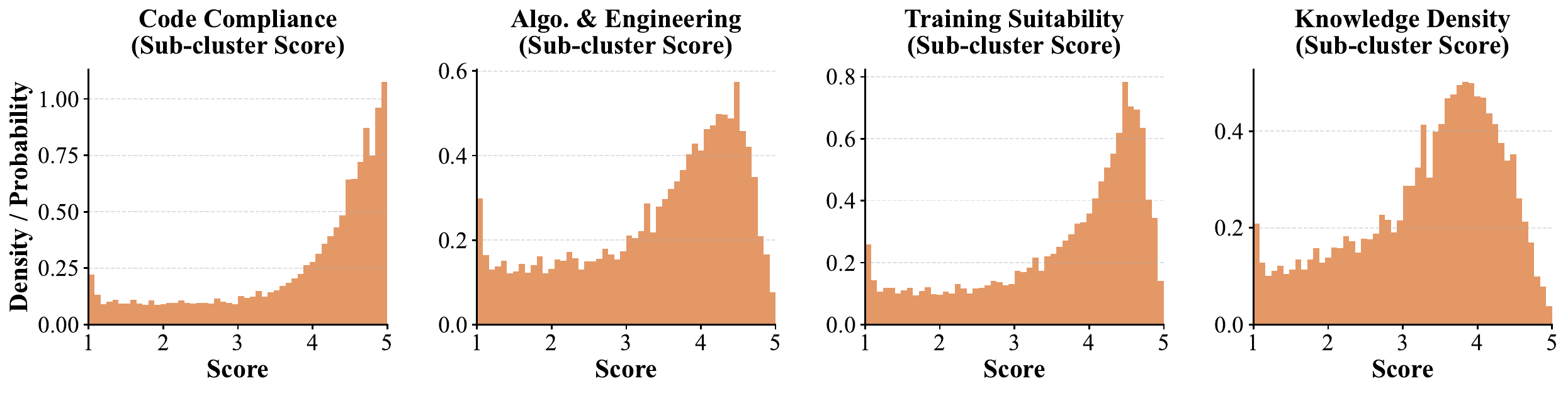}
    \vspace{-0.2cm}
    \caption{\textbf{Sub-cluster Semantic Score Distribution ($P_{S_j}$).} 
    Kernel density estimation of the aggregated semantic scores for sub-clusters $S_j$. The distribution shows a saturation effect in the $4.0-5.0$ range. This lack of variance in the top quantile underscores the necessity of incorporating geometric penalties ($\mathcal{L}_{struct}$) to introduce discriminative gradients among high-scoring clusters.}
    \label{fig:subcluster_density}
\end{figure*}

Figure~\ref{fig:subcluster_density} further illustrates the probability density of the aggregated Semantic Score ($P_{S_j}$) for sub-clusters. Consistent with the probe-level observations, the cluster-level scores are heavily concentrated in the high-value interval $[4.0, 5.0]$. This suggests that the majority of the code corpus, after initial filtering, is perceived as syntactically valid and high-quality by the LLM judge.

\subsection{Implications for Selection Strategy}
The observed "ceiling effect" in the score distributions highlights a potential limitation in relying solely on semantic scoring: the lack of discrimination in the high-score regime. 
Since a significant portion of sub-clusters achieves a saturated score ($P_{S_j} \approx 5.0$), semantic metrics alone may struggle to differentiate between intrinsic high-value domains and superficially correct samples. 
This empirical observation motivates the design of the \textbf{Structural } ($\mathcal{L}_{struct}$) within the UniGeM framework. By introducing geometric constraints as an orthogonal selection criterion, UniGeM effectively handles these saturated distributions, filtering out sub-clusters that appear high-quality but are topologically inconsistent with the domain manifold.

\cleardoublepage
\section{Reproducibility Statement}
\label{app:reproduct}

We provide sufficient details for an independent reimplementation of UniGeM and for reproducing our experimental results.

\textbf{Method / Algorithm.}
The full method specification (problem setup, Stage-I Macro-Exploration, Stage-II Micro-Mining, and all scoring terms) is described in Section~\ref{sec:method}. 

\textbf{Experimental Protocol.}
The evaluation protocol (corpus construction, sampling ratio, model settings, and compared baselines) is summarized in Section~\ref{sec:exp_setup}. Training and evaluation hyperparameters, together with the calibration logic for the geometric parameters, are provided in Appendix~\ref{sec:Experiment}.

\textbf{Baselines and Prompts.}
Details of the code adaptation for CLIMB and Meta-rater are documented in Appendix~\ref{sec:Experiment}. The probe/annotation rubric and the exact system prompt used for semantic scoring are provided in Appendix~\ref{sec:prompts}.

\textbf{Code Release.}
We will release the complete codebase (data processing, clustering/selection, training scripts, and evaluation) in a public repository upon completion of an internal review.

\textbf{Compute and infrastructure.}
All pre-training runs were executed on 64 NVIDIA H800 GPUs.
Each 100B-token pre-training run took approximately 15 hours wall-clock time.
Automated benchmark evaluation was executed on NVIDIA H20 GPUs.

\textbf{Use of AI assistants.}
AI assistants were used only for minor language polishing and clarity improvements during writing. All scientific content was developed and verified by the authors.

\end{document}